\documentclass[10pt,twocolumn,letterpaper]{article}

\usepackage[pagenumbers]{cvpr} 

\usepackage{graphicx}
\usepackage{amsmath}
\usepackage{amssymb}
\usepackage{booktabs}

\usepackage[table,xcdraw]{xcolor}
\usepackage{bm} 
\usepackage{pifont}
\newcommand{\cmark}{\ding{51}}
\newcommand{\xmark}{\ding{55}}
\usepackage{algorithm}
\usepackage{algpseudocode}
\usepackage{ctable}
\usepackage{array}
\usepackage{dcolumn}
\usepackage{multirow}
\usepackage[symbol]{footmisc}

\makeatletter
\def\@fnsymbol#1{\ensuremath{\ifcase#1\or \dagger\or \ddagger\or
   \mathsection\or \mathparagraph\or \|\or **\or \dagger\dagger
   \or \ddagger\ddagger \else\@ctrerr\fi}}
\makeatother

\DeclareMathOperator*{\argmin}{arg\,min}

\usepackage[pagebackref,breaklinks,colorlinks]{hyperref}

\usepackage[capitalize]{cleveref}
\crefname{section}{Sec.}{Secs.}
\Crefname{section}{Section}{Sections}
\Crefname{table}{Table}{Tables}
\crefname{table}{Tab.}{Tabs.}

\def\cvprPaperID{8822} 
\def\confName{CVPR}
\def\confYear{2023}

\begin{document}

\title{BlackVIP: Black-Box Visual Prompting for Robust Transfer Learning}

\author{Changdae Oh$^1$ \;\, Hyeji Hwang$^1$ \;\, Hee-young Lee$^2$\thanks{Work done at University of Seoul} \;\, YongTaek Lim$^1$\;\, Geunyoung Jung$^1$ \\ Jiyoung Jung$^1$\;\, Hosik Choi$^1$\;\, Kyungwoo Song$^3$\thanks{Corresponding author; Work partly done at University of Seoul} \\
$^1$University of Seoul \,\; $^2$Sungkyunkwan University \,\; $^3$Yonsei University\\
{\tt\small changdae.oh@uos.ac.kr \;\;\; kyungwoo.song@yonsei.ac.kr}}
\maketitle

\begin{abstract}
With the surge of large-scale pre-trained models (PTMs), fine-tuning these models to numerous downstream tasks becomes a crucial problem. Consequently, parameter efficient transfer learning (PETL) of large models has grasped huge attention. While recent PETL methods showcase impressive performance, they rely on optimistic assumptions: 1) the entire parameter set of a PTM is available, and 2) a sufficiently large memory capacity for the fine-tuning is equipped. However, in most real-world applications, PTMs are served as a black-box API or proprietary software without explicit parameter accessibility. Besides, it is hard to meet a large memory requirement for modern PTMs. In this work, we propose black-box visual prompting (BlackVIP), which efficiently adapts the PTMs without knowledge about model architectures and parameters. BlackVIP has two components; 1) Coordinator and 2) simultaneous perturbation stochastic approximation with gradient correction (SPSA-GC). The Coordinator designs input-dependent image-shaped visual prompts, which improves few-shot adaptation and robustness on distribution/location shift. SPSA-GC efficiently estimates the gradient of a target model to update Coordinator. Extensive experiments on 16 datasets demonstrate that BlackVIP enables robust adaptation to diverse domains without accessing PTMs' parameters, with minimal memory requirements. Code: \url{https://github.com/changdaeoh/BlackVIP}
\end{abstract}


\section{Introduction}
\label{sec:intro}
\begin{figure}[t]
    \centerline{\includegraphics[width=\linewidth]{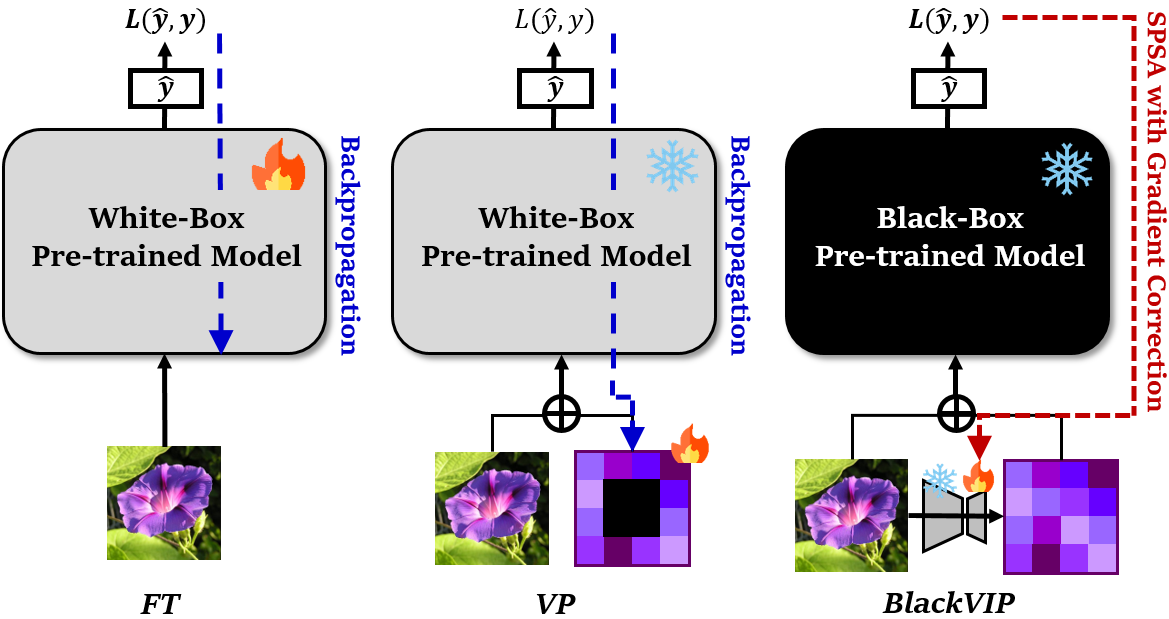}}
    \caption{While FT updates the entire model, VP has a small number of parameters in the input pixel space. However, VP still requires a large memory capacity to optimize the parameters through backpropagation. Moreover, FT and VP are only feasible if the PTM's parameters are accessible. Meanwhile, BlackVIP does not assume the parameter-accessibility by adopting a black-box optimization (SPSA-GC) algorithm rather than relying on backpropagation. Besides, BlackVIP reparameterizes the visual prompt with a neural network and optimizes tiny parameters with SPSA-GC. Based on the above properties, BlackVIP can be widely adopted in realistic and resource-limited transfer learning scenarios.}
	\label{fig:overview_illustration}
\end{figure}

Based on their excellent transferability, large-scale pre-trained models (PTMs) \cite{brown2020language, dosovitskiy2020image, radford2021learning} have shown remarkable success on tasks from diverse domains and absorbed increasing attention in machine learning communities. By witnessing PTMs' success, Parameter-Efficient Transfer Learning (PETL) methods that efficiently utilize the PTMs are recently emerging. While the standard fine-tuning (FT) and its advanced variants \cite{wortsman2022robust, kumar2021fine} update the entire or large portion of a PTM \cite{devlin2018bert}, PETL methods aim to achieve comparable performance to FT by optimizing a small number of learnable parameters. 

Among them, \textit{prompt-based approaches} \cite{li2021prefix, lester2021power, jia2022visual, bahng2022visual, bar2022visual} have been widely investigated from diverse research areas. For vision PTMs, Visual Prompt Tuning \cite{jia2022visual} injects a few additional learnable prompt tokens inside of ViT's \cite{dosovitskiy2020image} layers or embedding layer and only optimizes them. Bahng et al. \cite{bahng2022visual} investigate visual prompting (VP), which adopts the learnable parameters on input pixel space as a visual prompt, while no additional modules are inserted into the pre-trained visual model. Besides, prompt learning methods for VLM are also actively studied \cite{ju2021prompting, zhou2022learning, zhou2022conditional, zang2022unified}.

While existing PETL methods show impressive performance with few learnable parameters, they rely on two optimistic assumptions. First, the previous PETL assumes that the full parameters of the PTM are accessible. However, many real-world AI applications are served as API and proprietary software, and they do not reveal the implementation-level information or full parameters due to commercial issues, e.g., violating model ownership. As a result, exploiting high-performing PTMs to specific downstream tasks not only in the white-box setting but also black-box setting (limited accessibility to the model's detail) is a crucial but unexplored problem. Second, existing methods require a large memory capacity. While PETL approaches have few learnable parameters, they require a large amount of memory for backpropagating the gradient throughout the large-scale PTM parameters to learnable parameters. Therefore, users who want to adopt a large-scale PTM should satisfy large memory requirements despite the small learnable parameters. Besides, if the users entrust PTM fine-tuning to the model owner with their specific data, data-privacy concerns will inevitably arise \cite{xiao2023offsite}.

To alleviate the above unrealistic assumptions, we are pioneering \textbf{\textit{black-box visual prompting} (BlackVIP)} approach, which enables the parameter-efficient transfer learning of pre-trained black-box vision models from the low-resource user perspective (illustrated in Figure \ref{fig:overview_illustration}). BlackVIP works based on the following two core components: 1) pixel space input-dependent visual prompting and 2) a stable zeroth-order optimization algorithm.

Firstly, we augment an input image by attaching an visual prompt per pixel. It is noted that input space prompting does not require the accessibility on parts of architecture \cite{khattak2022maple, zang2022unified} or the first embedding layer \cite{zhou2022learning, zhou2022conditional, ju2021prompting} of PTM. While the previous works only introduce a pixel-level prompt to a small fraction of the fixed area, such as outside of the image \cite{bahng2022visual}, BlackVIP designs the prompt with the same shape as the original given image to cover the entire image view. Therefore, our prompt has a higher capability and can flexibly change the semantics of the original image. In addition, we reparameterize the prompt with a neural network. Specifically, we propose the \textbf{\textit{Coordinator}}, an asymmetric autoencoder-style network that receives the original image and produces a corresponding visual prompt for each individual image. As a result, Coordinator automatically designs each prompt conditioned on the input rather than the shared manual design of a previous work \cite{bahng2022visual}. By optimizing the reparameterized model instead of the prompt itself, we greatly reduce the number of parameters (from 69K of VP \cite{bahng2022visual} to 9K) so that suitable for black-box optimization. 

Next, unlike other PETL approaches, BlackVIP adopts a zeroth-order optimization (ZOO) that estimates the zeroth-order gradient for the coordinator update to relax the assumption that requires access to the huge PTM parameters to optimize the prompt via backpropagation. Therefore, BlackVIP significantly reduces the required memory for fine-tuning. Besides, we present a new ZOO algorithm, \textbf{\textit{Simultaneous Perturbation Stochastic Approximation with Gradient Correction} (SPSA-GC)} based on (SPSA) \cite{119632}. SPSA-GC first estimates the gradient of the target black-box model based on the output difference of perturbed parameters and then corrects the initial estimates in a momentum-based look-ahead manner. By integrating the Coordinator and SPSA-GC, BlackVIP achieves significant performance improvement over baselines.

Our main contributions are summarized as follows:
\begin{itemize}
  \item To our best knowledge, this is the first paper that explores the input-dependent visual prompting on black-box settings. For this, we devise Coordinator, which reparameterizes the prompt as an autoencoder to handle the input-dependent prompt with tiny parameters.
  \item We propose a new ZOO algorithm, SPSA-GC, that gives look-ahead corrections to the SPSA's estimated gradient resulting in boosted performance. 
  \item Based on Coordinator and SPSA-GC, BlackVIP adapts the PTM to downstream tasks without parameter access and large memory capacity. We extensively validate BlackVIP on 16 datasets and demonstrate its effectiveness regarding few-shot adaptability and robustness on distribution/object-location shift.
\end{itemize}

\section{Related Works}
\label{sec:rel_work}

\subsection{Pre-trained Vision Models}
Over the past decade, the pre-train and fine-tune paradigm has become the de-facto standard using deep neural networks. Beyond the label supervision \cite{simonyan2014very, he2016deep}, self-supervised learning (SSL) \cite{he2020momentum, chen2020simple, grill2020bootstrap, zbontar2021barlow, he2022masked, xie2022simmim} approaches that do not rely on human-annotated labels hit the machine learning community. SSL approaches can roughly be categorized into discriminative and generative approaches. Discriminative SSL methods \cite{he2020momentum, chen2020simple, grill2020bootstrap, zbontar2021barlow} learn the embeddings by enforcing closeness and/or distantness on the pair-wise distance structure among the augmented training samples. Meanwhile, the recently emerging generative SSL methods \cite{bao2021beit, he2022masked, xie2022simmim} are based on \textit{masked image modeling}, which supervises the model by encoding and reconstructing the partially masked individual images. SSL approaches are appealing not only due to their label-free training regime but also produce a more transferable representation \cite{caron2021emerging, he2022masked, liu2022selfsupervised} getting over the pre-defined label category.

Moreover, fuelled by pre-training with rich semantic structures from image-caption pairs of the web-scale dataset, visual-language pre-trained models \cite{radford2021learning, jia2021scaling, yuan2021florence, singh2022flava} recently showed surprising performance on the zero-shot transfer and few-shot adaptation. Based on their high transferability, they are being adopted for numerous downstream tasks from diverse domains. Meanwhile, the number of parameters of PTMs has increased continuously, showing the performance improvement proportional to the number of parameters \cite{kaplan2020scaling}. However, large models require sufficiently large memory capacity in the fine-tuning stage. Besides, it is commonly impossible to access the PTMs' parameters in public. Therefore, we propose a new fine-tuning method that does not require both knowledge about model parameters and a large amount of memory.

\subsection{Parameter-Efficient Transfer Learning}
To adapt the large-scale PTMs to targeted downstream tasks, Parameter-Efficient Transfer Learning (PETL) methods pursue fine-tuning of a small subset of large PTMs, while achieving competitive performance compared to full fine-tuning. Recently, diverse PETL approaches have emerged in the NLP domain, such as adapter \cite{houlsby2019parameter, pfeiffer2020adapterfusion} and prompt learning \cite{li2021prefix, lester2021power}.

Motivated by the promising results of PETL in NLP, there have been many efforts to realize PETL in vision or vision-language fields. For the case of adapter-based methods, AdaptFormer \cite{chen2022adaptformer}, and  CLIP-Adapter \cite{gao2021clip} insert a few learnable modules inside of the vision encoder (e.g., ViT \cite{dosovitskiy2020image}) or on top of both the vision and text encoder, respectively. In the case of prompt-based approaches, CoOp \cite{zhou2022learning} introduces the continuous text prompt into the text encoder of a VLM, and Conditional CoOp (CoCoOp) \cite{zhou2022conditional} extends CoOp to an input-dependent version. Besides, Jia et al. \cite{jia2022visual} propose the Visual Prompt Tuning (VPT) that governs learnable visual tokens to the embedding layer (VPT-Shallow) or several encoder layers (VPT-Deep) of ViT. Bahng et al. \cite{bahng2022visual} explore the Visual Prompting (VP) approach, which introduces a learnable prompt in the input space, not into the embedding space or model's building blocks. Then the prompt is attached to the image in the fixed restricted region. Among them, VP is especially attractive because it does not require full accessibility of the model architecture and parameters during the inference phase, which motivates us to investigate the \textit{black-box visual prompting}.

However, we argue that the existing visual prompt approaches can be advanced from three perspectives. (1) \textit{From white-box to black-box}: to be utilized for a variety of real-world applications, PETL should be able to deal with black-box PTMs that are served via API or proprietary software; for this, black-box optimization methods are required rather than backpropagation in previous works. (2) \textit{Towards input-dependent visual prompt}: the visual features of individual images are distinct even though the images share the same class label; therefore, the input-dependent prompt is necessary. (3) \textit{beyond the manual prompt design}: the prompt of VP is manually designed like a frame- or square-shaped and attached to a restricted region; this limited flexibility induces a sub-optimal prompt on challenging generalization scenario (refer to the Sec \ref{sec:toy}). To this end, we propose BlackVIP, which adopts ZOO rather than backpropagation and automatically designs input-dependent prompts over the entire image region. Table \ref{tab:related} summarizes the comparison between the previous methods and ours.
 
\begin{table}[htbp]
\centering
\scriptsize
\caption{Prompt-based PETL. Loc. denotes the prompt location.}
\vspace{-0.5em}
\begin{tabular}{@{}lccccc@{}}
\toprule
Method   & Grad-free & Prompt & Loc. & Input-dependent \\ \toprule
CLIP ZS \cite{radford2021learning}     & {\color[HTML]{17C211} \cmark} &L& \cellcolor[HTML]{FFFC9E}\textit{input } & {\color[HTML]{E80F0F} \xmark} \\ 
CoOp \cite{zhou2022learning}    & {\color[HTML]{E80F0F} \xmark} &L& \cellcolor[HTML]{C0C0C0}\textit{emb } & {\color[HTML]{E80F0F} \xmark} \\ 
CoCoOp \cite{zhou2022conditional}  & {\color[HTML]{E80F0F} \xmark} &L& \cellcolor[HTML]{C0C0C0}\textit{emb } & {\color[HTML]{17C211}\cmark} \\
VPT \cite{jia2022visual}     & {\color[HTML]{E80F0F} \xmark} &V& \cellcolor[HTML]{C0C0C0}\textit{emb } & {\color[HTML]{E80F0F} \xmark} \\
VP \cite{bahng2022visual}      & {\color[HTML]{E80F0F} \xmark } &V& \cellcolor[HTML]{FFFC9E}\textit{input } & {\color[HTML]{E80F0F} \xmark} \\
BAR \cite{tsai2020transfer} & {\color[HTML]{17C211} \cmark} &V& \cellcolor[HTML]{FFFC9E}\textit{input } & {\color[HTML]{E80F0F}\xmark} \\
BlackVIP (Ours) & {\color[HTML]{17C211} \cmark} &V& \cellcolor[HTML]{FFFC9E}\textit{input } & {\color[HTML]{17C211}\cmark} \\ \bottomrule
\end{tabular}
\label{tab:related}
\vspace{-0.5em}
\end{table}

\subsection{Black-Box Optimization}
Numerous high-performing artificial intelligence models have been deployed, and many custom services are based on the API or proprietary software. There are several works on NLP field that fine-tune the large language model via black-box optimization \cite{sun2022bbt, sun2022bbtv2, deng2022rlprompt}. Besides, black-box adversarial reprogramming (BAR) \cite{tsai2020transfer} had been proposed to re-purpose the ImageNet \cite{5206848} pre-trained vision model to a medical image classifier.

The previous works on black-box attack and optimization utilize ZOO algorithms or derivative-free optimization algorithms for parameter updates. BAR \cite{tsai2020transfer} adopts a one-sided approximation gradient estimator, but we find that the one-sided estimator shows inaccurate gradient approximations empirically. BBT \cite{sun2022bbt}, and BBTv2 \cite{sun2022bbtv2} adopt Covariance Matrix Adaptation Evolution Strategy (CMA-ES) \cite{hansen2001completely, hansen2003reducing}, and RLPrompt \cite{deng2022rlprompt} uses reinforcement learning (Soft Q-Learning \cite{guo2021text}) to optimize the discrete prompts. However, It has been known that derivative-free optimizations (e.g. evolutionary optimization) are hard to solve large-scale problems and do not guarantee convergence \cite{liu2020primer}. Besides, reinforcement learning algorithms are notorious for their unstable optimization, and high variance \cite{zhao2011analysis}. 

In this work, we adopt the Simultaneous Perturbation Stochastic Approximation (SPSA) \cite{119632} as a ZOO algorithm. It is known that SPSA is efficient at high-dimensional gradient approximation problems \cite{119632, spall1998overview}. Besides, SPSA theoretically guarantees convergence, and the convergence error is linearly upper bounded by the parameter dimension \cite{119632}. While SPSA is designed to estimate high-dimensional gradients efficiently, we found that SPSA-based neural network optimization still requires many queries in practice. Therefore, we propose SPSA with Gradient Correction (SPSA-GC) that corrects the approximated gradients to enhance the convergence speed. To our best knowledge, this is the first work exploring the ZOO-based black-box optimization to large PTMs for general-purpose adaptation (rather than a specific domain \cite{tsai2020transfer}).

\begin{figure*}[!t]
    \centerline{\includegraphics[width=1.0\textwidth]{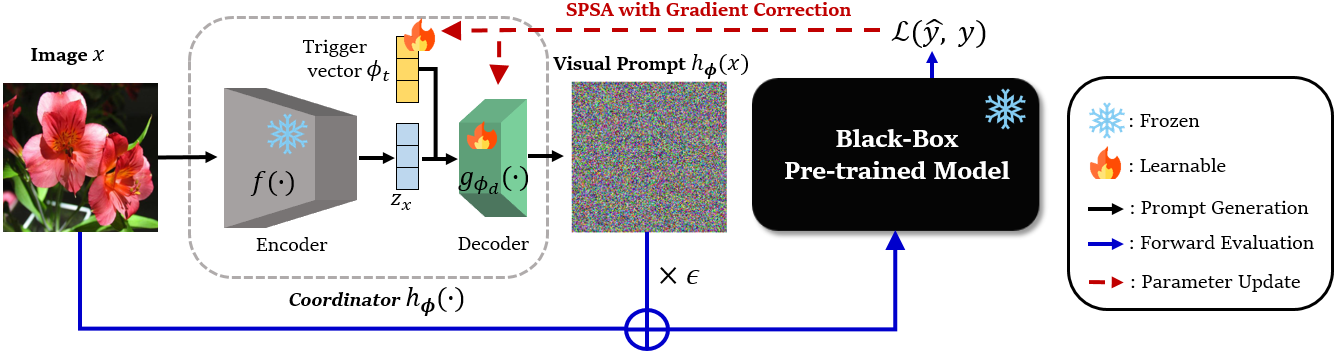}}
    \caption{BlackVIP equips an input-dependent prompt designer (Coordinator) and an accurate gradient estimation algorithm (SPSA-GC).}
	\label{fig:BlackVIP_framework}
 \vspace{-1em}
\end{figure*}

\section{Preliminary}

We first present an outline of \textit{adversarial reprogramming} and \textit{visual prompting}, that originated from distinct motivations, but closely related topics. Elasyed et al. \cite{elsayed2018adversarial} presented \textit{adversarial reprogramming} (AR) inspired by adversarial attack \cite{goodfellow2014explaining, 7467366, xu2020adversarial}. The goal of AR is repurposing the pre-trained model to perform a new task.
Let $x \in \mathbb{R}^{k \times k \times 3}$ be a downsized image from the adversarial target dataset, and $\Tilde{x} \in \mathbb{R}^{n \times n \times 3}$ is a random image from pre-train dataset or a zero-padded tensor that includes $x$ in the center of image $\Tilde{x}$, where $k<n$. Given the target class of adversarial task $y_{adv} \in \{1, ..., C_{tar}\}$, the AR is formulated as:
\begin{align}
	\argmin_{W} (-\log{P_{\theta;W}(h(y_{adv})|\Tilde{x}_{adv}}) + ||W||_{F}) \nonumber
\end{align}
Here, $\theta$ is the pre-trained model parameters, and the adversarial image is constructed as $\Tilde{x}_{adv}=\Tilde{x}+\text{tanh}(W \odot M)$, and $W \in \mathbb{R}^{n \times n \times 3}$ is the adversarial program that is optimized, where $n$ is the image width of a pre-train dataset, $M$ is an optional mask for better visualization of the embedded target image, and $\odot$ denotes the element-wise multiplication. Given a pre-defined hard-coded mapping $h(\cdot)$ that maps labels from an adversarial task to labels of a pre-train dataset, AR reprograms the model via learned perturbation without architectural change. The vulnerability of neural networks to adversarial examples has inspired many works that use the AR approach from transfer learning perspectives \cite{tsai2020transfer, neekhara2022cross, chen2022model, melnyk2022reprogramming} i.e., \textit{model reprogramming}.

Meanwhile, motivated by the remarkable success of the \textit{prompting} paradigm on NLP, Bahng et al. \cite{bahng2022visual} are the first to explore the input pixel space \textit{visual prompting} (VP) approach for pre-trained vision and vision-language models. By learning pixel-style prompts (i.e., perturbation) attached to the input images, VP adapts the frozen PTM to targeted downstream tasks without modifying on model architecture. Given the input image $x$ and corresponding label $y$, the learning objective of VP is as follows: 
\begin{align}
	\argmin_{\phi} -\log P_{\theta;\phi}(y|x+\phi) \nonumber
\end{align}
where $\theta$ and $\phi$ are the PTM parameters and visual prompt, respectively. At the inference phase, VP employs the shared prompt (input-independent) for all images. It is noted that $\phi$ is attached to the fixed location, e.g., the outer part of the image like a frame by default. 

Though AR and VP use different terms and stem from distinct motivations, they share the general idea: adapt a PTM to perform new tasks without modifying the model architecture. This paper aligns with AR and VP, but broadens and improves them for more realistic environments.

\section{Methodology}
We introduce our novel input-dependent prompt generation module, \textit{Coordinator} (in Section \ref{sec:method_reparam}). Then, we explain the end-to-end framework of BlackVIP with the new ZOO algorithm, \textit{SPSA-GC} (in Section \ref{sec:method_spsa}). Figure \ref{fig:BlackVIP_framework} illustrates the overall framework of BlackVIP.
\subsection{Coordinator: Prompt Reparameterization}
\label{sec:method_reparam}
Our learning objective is to minimize the downstream task loss by adapting the frozen PTM via input space prompt optimization. Given a frozen prediction model $P_{\theta}(y|x)$, and perturbed image $\Tilde{x}$ with prompt corresponding to the label $y$, the training objective is formulated as:
\begin{align}
\argmin_{\bm\phi} - \log P_{\theta;\bm\phi}(y|\Tilde{x}) 
\nonumber
\end{align}
While VP and AR optimize the input space visual prompt directly, we reparameterize the visual prompt to the prompt generation network $h_{\bm\phi}(\cdot)$ parameterized by $\bm\phi=\{ \phi_d, \phi_t\}\in \mathbb{R}^{d}$. Specifically, we build a novel autoencoder-style network named Coordinator composed of a frozen encoder $f(\cdot)$ which is pre-trained on ImageNet \cite{5206848} by self-supervised learning (SSL) objective and followed by an extremely light-weight learnable decoder $g_{\phi_d}(\cdot)$. Though the encoder can also be a supervised counterpart or light-weight learnable network, we adopt the SSL pre-trained encoder for the following three reasons: 1) It has been widely substantiated that self-supervised representation contains the multiple discriminative features and spatial information \cite{caron2021emerging, he2022masked, li2021benchmarking, liu2022selfsupervised, huang2022survey, fang2022unleashing, pan2022towards}, so it is more helpful to use SSL pre-trained encoder than label-supervised encoder for robustly performing on diverse downstream tasks. 2) ImageNet pre-trained encoders are currently well-popularized \cite{tensorflow2015-whitepaper, NEURIPS2019_9015, rw2019timm, wolf-etal-2020-transformers}, so they can be easily adopted by local users, and does not hurt our realistic experimental setting. 3) By using the frozen pre-trained encoder, we significantly reduce the number of learnable parameters. The reduced low-dimensional parameters encourage efficient gradient approximation. Consequently, the image equipped with a prompt (prompted image) is constructed as follows:
\begin{align}
\Tilde{x} &= \text{clip}(x+\epsilon h_{\bm\phi}(x)) \nonumber \\
h_{\bm\phi}(x) &= g_{\phi_{d}}(z_x, \phi_t) \nonumber 
\label{eq:prompted_image}
\end{align}
\noindent where $z_{x}=f(x)$ is the feature vector of $x$ from the frozen SSL encoder $f(\cdot)$, and $\epsilon \in [0,1]$ is a hyperparameter that controls the intensity of visual prompt. Here, $\phi_{t}$ is a task-specific \textit{prompt trigger vector} that is jointly optimized with decoder parameter $\phi_{d}$. We concatenate it with $z_{x}$ and then reshape them into a 3D feature map to feed into the convolutional decoder. As a result, the instance-specific rich semantic representation $z_x$ and the task-specific prompt trigger vector $\phi_{t}$ are merged to design a valid visual prompt $h_{\phi}(x)$ for a given image. Similar to \cite{neekhara2022cross}, the prompted image is bounded to a valid RGB scale via pixel-wise clipping. 

Unlike previous visual prompts (e.g., VP) or adversarial programs (e.g., BAR), our BlackVIP automatically designs the input-dependent prompts with the same shape as the original images; therefore, it has a higher capability to change the semantics of images if necessary. Thanks to this flexibility, BlackVIP can cover more diverse tasks and be robust to challenging scenarios, e.g., distribution shift.
\subsection{End-to-End Black-Box Visual Prompting}
\label{sec:method_spsa}
Unlike other PETL approaches that assume the accessibility to the architecture and/or parameters of the PTM, we consider the PTM as a black-box predictor that gives only a prediction output (i.e. logit) for a given input image query. In this black-box setting, we adopt the ZOO algorithm, SPSA, with our considerate modification to optimize our Coordinator without the oracle true gradient.
\paragraph{SPSA} Spall et al. proposed Simultaneous Perturbation Stochastic Approximation (SPSA) \cite{119632, SPALL1997109} that approximates the high-dimensional gradient efficiently. Given the positive decaying sequences of $a_{i}>0$ and $c_{i}\in [0,1]$, the gradient approximation, $\hat{g}$, and single-step parameter update of SPSA is described as follows:
\begin{align}
\hat{g}_{i}(\bm\phi_{i}) &= {L(\bm\phi_{i} + c_i\Delta_{i}) - L(\bm\phi_{i} - c_i\Delta_{i}) \over 2 c_i} \Delta_{i}^{-1}  \\
\bm\phi_{i+1} &= \bm\phi_{i} - a_i \hat{g}_{i}(\bm\phi_{i})
\label{eq:spsa}
\end{align}
\noindent where $L$ is an objective function, $\bm\phi_{i}\in \mathbb{R}^{d}$ is d-dimensional learnable parameters, and $\Delta_{i} \in \mathbb{R}^d$ is a $i^{th}$-step random perturbation vector, sampled from mean-zero distributions that satisfy finite inverse momentum condition \cite{119632, 10.5555/773301} such as Rademacher and Segmented Uniform distribution. With only two forward evaluations, i.e., querying twice to the API service model, SPSA parses the learning signal (estimate gradient) from the model's output difference, and we can optimize the parameters of Coordinator $\bm\phi$ to design the proper visual prompt for a given input.

\paragraph{SPSA with Gradient Correction} Although the standard form of SPSA works well in myriad applications \cite{4469948, stein13_interspeech, 9143831}, like other ZOO algorithms, it may suffer slow convergence in practice \cite{880982, Spall1997AcceleratedSS}, and the problem gets even bigger on the high-dimensional problem setting such as neural networks' optimization. We speculate that the source of slow convergence is its noisy gradient estimation from the poor direction of random perturbations or intrinsic data noise. To mitigate this estimation noise, inspired by Nesterov's accelerated gradient (NAG) \cite{Nesterov1983AMF}, we improve the parameter update rule in Eq. \ref{eq:spsa} as below:
\begin{align}
\bm\phi_{i+1} &= \bm\phi_{i} + m_{i+1} \\ \nonumber
m_{i+1} &= \beta m_{i} - a_{i}\hat{g}_{i}(\bm\phi_{i} + \beta m_{i}) \nonumber
\label{eq:spsa_gc}
\end{align}
\noindent where $\beta \in [0,1]$ is smoothing parameter. As clearly noted in \cite{pmlr-v28-sutskever13}, when the poor update $\bm\phi_{i} + \beta m_{i}$ occurs, this NAG style update rule strongly pulls it back towards $\bm\phi_{i}$. Because of the SPSA's strongly stochastic nature, we conjecture that this \textit{gradient correction} property is also highly effective for SPSA as well as first-order optimization algorithms. Algorithm \ref{alg:BlackVIP} summarizes the BlackVIP algorithm.
\begin{algorithm}[t]
\caption{BlackVIP algorithm}
\label{alg:BlackVIP}
\begin{algorithmic}
    \Require{
    Downstream dataset $\mathcal{D}$, pre-trained model $P_{\theta}$, Coordinator $h$ with encoder $f$  and prompt decoder $g$, is parameterized by $\bm\phi_{i}=\{\phi_{d,i}, \phi_{t,i}\}$, SPSA-GC decaying parameters $\{ a_{i},c_{i} \}$, and smoothing parameter $\beta$, prompt intensity $\epsilon$, and training iteration $R$.}
    \State \textcolor{gray}{// Initialize $\bm\phi_{1}=\{\phi_{d,1}, \phi_{t,1}\}$, $\{a_{1}, c_{1}\}$ and $m_{1}$ }
    \For{$i$ in $1$ to $R$}
    \State \textcolor{gray}{// Parse a batch $(x, y) \sim \mathcal{D}$ and design the prompt}
    \State $h_{\bm\phi_{i}}(x) = g_{\phi_{d,i}}(f(x), \phi_{t,i})$
    \State $\Tilde{x} = \text{clip}(x+\epsilon h_{\bm\phi_{i}}(x))$
    \State \textcolor{gray}{// Draw a sample $\Delta_{i}$, set $c_{i}$, and estimate the gradient}
    \State $L(\bm\phi_{i}) :=  - \log P_{\theta;\bm\phi_{i}}(y|\Tilde{x})$
    \State $\hat{g}_{i}(\bm\phi_{i}) = (L(\bm\phi_{i} + c_i\Delta_{i}) - L(\bm\phi_{i} - c_i\Delta_{i})) (2c_{i}\Delta_{i})^{-1}$
    \State \textcolor{gray}{// Set $a_{i}$, and update parameters}
    \State $m_{i+1} = \beta m_{i} - a_{i}\hat{g}_{i}(\bm\phi_{i} + \beta m_{i})$
    \State $\bm\phi_{i+1} = \bm\phi_{i} + m_{i+1} $
    \EndFor 
\end{algorithmic}
\end{algorithm}

\section{Results}
We first provide the experimental setup in Section \ref{sec:exp_setup}. Next, Section \ref{sec:toy} presents the comparison between SPSA-GC and the previous ZO method. Besides, we provide domain generalization and object location sensitivity experiments. Section \ref{sec:few} and \ref{sec:ablation} provide the results on 14 transfer learning benchmarks and ablation studies, respectively.

\subsection{Experimental Setup} \label{sec:exp_setup}
\begin{figure}[htbp]
     \centering
     \includegraphics[width=\linewidth]{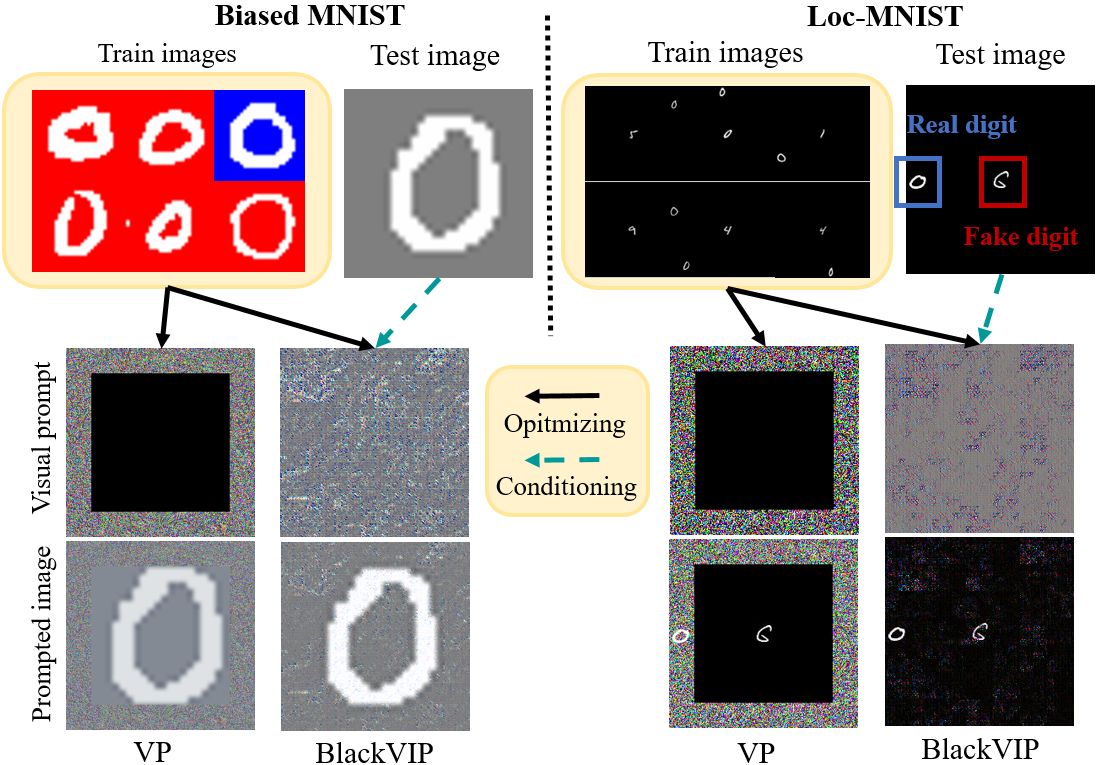} %
    \caption{Prompt visualization on synthetic datasets. Unlike VP, our BlackVIP designs input-dependent conditional prompts contributing to the robustness under distribution/object-location shift.}
    \label{fig:synthetic_illu}
\vspace{-1em}
\end{figure}

We extensively evaluate BlackVIP on 14 benchmarks (refer Supp A.1). These cover diverse visual domains and tasks, so they require understanding various visual semantics like scenes, actions, fine-grained categories, textures, satellite imagery, the number of objects, and the recognition of generic objects. Additionally, to investigate the importance of prompt design, we consider two synthetic datasets: Biased MNIST and Loc-MNIST (see Sec \ref{sec:toy} and Fig. \ref{fig:synthetic_illu}).

In this paper, we adopt CLIP ViT-B/16 \cite{radford2021learning} as a target PTM because it does not require a separate classifier for different tasks, and has a strong zero-shot generalization capability. For the frozen encoder of Coordinator, we use ImageNet pre-trained \texttt{vit-mae-base} checkpoint.
As the baselines, we consider CLIP's zero-shot classifier (ZS), black-box adversarial reprogramming (BAR) \cite{tsai2020transfer}, and VP with SPSA-GC that simply replace the backpropagation in VP \cite{bahng2022visual} with SPSA-GC. Following the few-shot classification setting of \cite{zhou2022learning}, we use 16-shot training samples and the full testset by default. More details are provided in Supp A.

\subsection{Synthetic Datasets} \label{sec:toy}
\begin{figure}[h!]
     \centering
     \begin{subfigure}[b]{0.485\linewidth}
         \centering
        \includegraphics[width=0.95\linewidth]{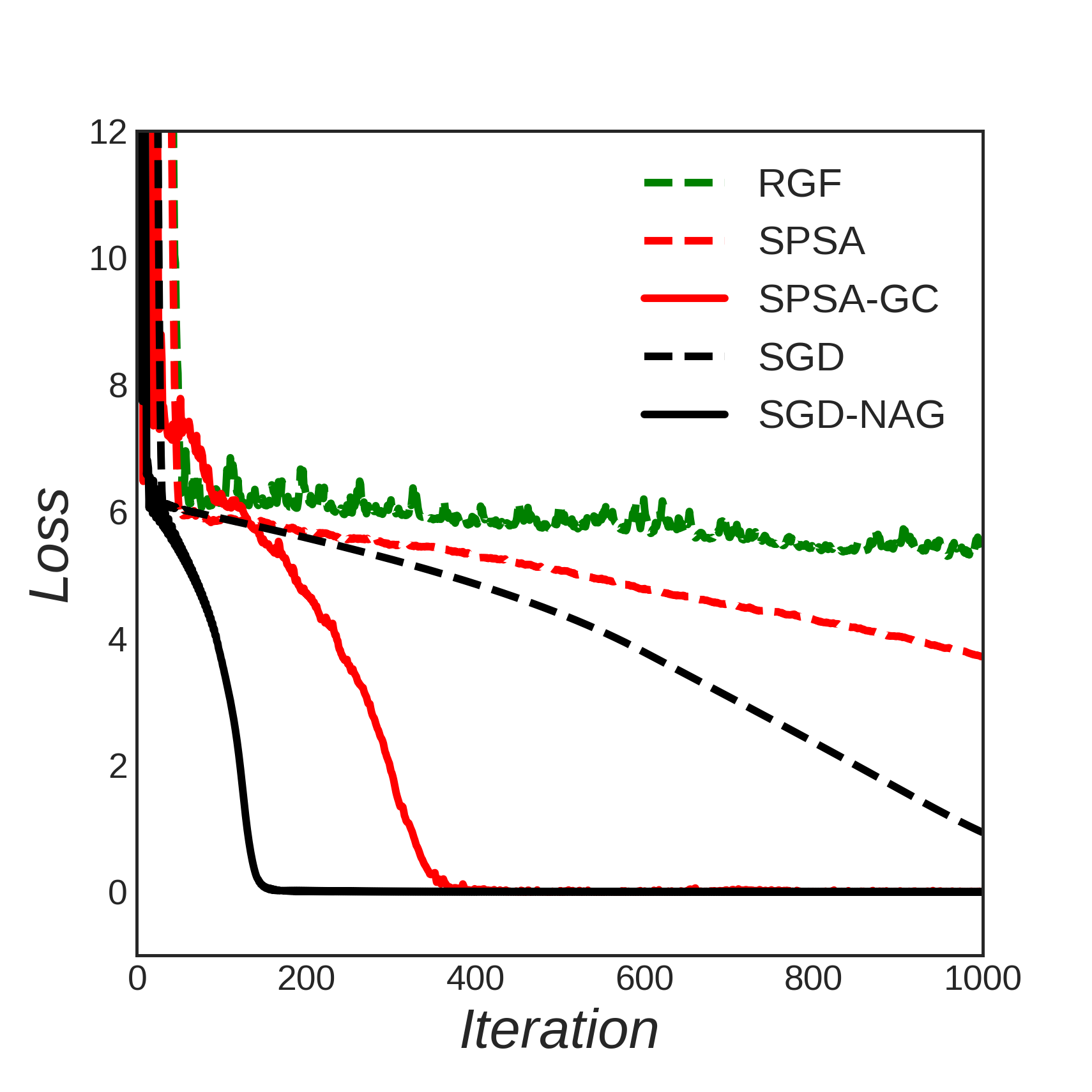}
     \end{subfigure}
     \hfill
     \begin{subfigure}[b]{0.48\linewidth}
         \centering
        \includegraphics[width=0.90\linewidth]{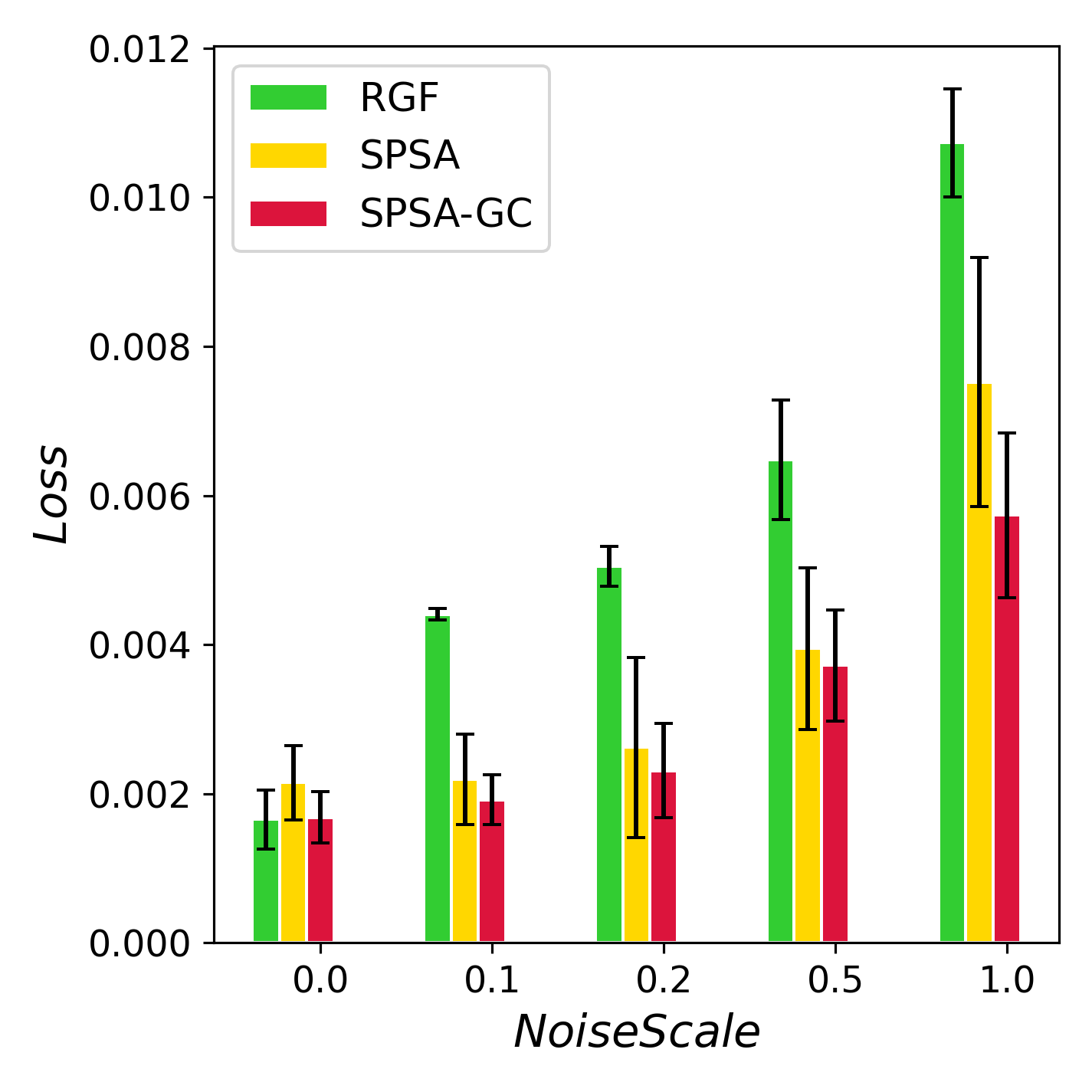}
     \end{subfigure}
    \caption{(Left) loss curve and (right) noise sensitivity analysis of 100-Dimensional Rosenbrock optimization.}
\label{fig:opt}
\vspace{-1em}
\end{figure}

\paragraph{Comparison among optimization algorithms}
We validate our SPSA-GC on the well-known optimization benchmark, Rosenbrock function. We report the normalized loss (${|L(\theta^{*})-L(\theta)| \over |L(\theta^{*})-L(\theta_{0})|}$) where $L(\theta^{*})$ and $L(\theta_{0})$ is the loss value on the optimal and initial point, respectively, and $L(\theta)$ is a loss value on the current parameter $\theta \in \mathbb{R}^{100}$. In Fig. \ref{fig:opt} (left), SPSA-GC shows faster and more stable convergence than Random Gradient-Free (RGF) \cite{liu2018zeroth, tsai2020transfer}, and even achieves a comparable result to Nesterov's Accelerated Gradient (SGD-NAG) using true gradients. Besides, we simulate the noisy loss observation (emulating the mini-batch optimization) by adding Gaussian noise to learning loss, i.e., $L_{noisy}(\theta)=L(\theta)+\epsilon$, where $\epsilon \sim N(0,scale^{2})$. In Fig. \ref{fig:opt} (right), as the noise increases, RGF rapidly degenerates while SPSA is still relatively stable, and our gradient correction (SPSA-GC) gives further improvement.

\begin{figure}[b]
\vspace{-1em}
     \centering
     \begin{subfigure}[t]{0.485\linewidth}
         \centering
        \includegraphics[width=0.95\linewidth]{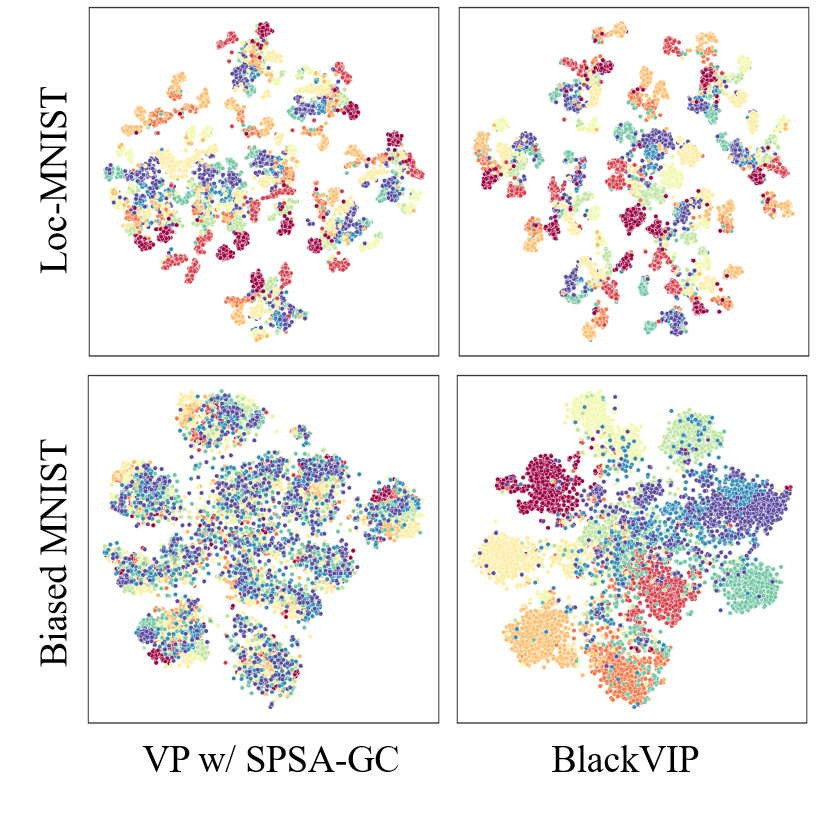}
     \end{subfigure}
     \hfill
     \begin{subfigure}[t]{0.485\linewidth}
         \centering
        \includegraphics[width=0.98\linewidth]{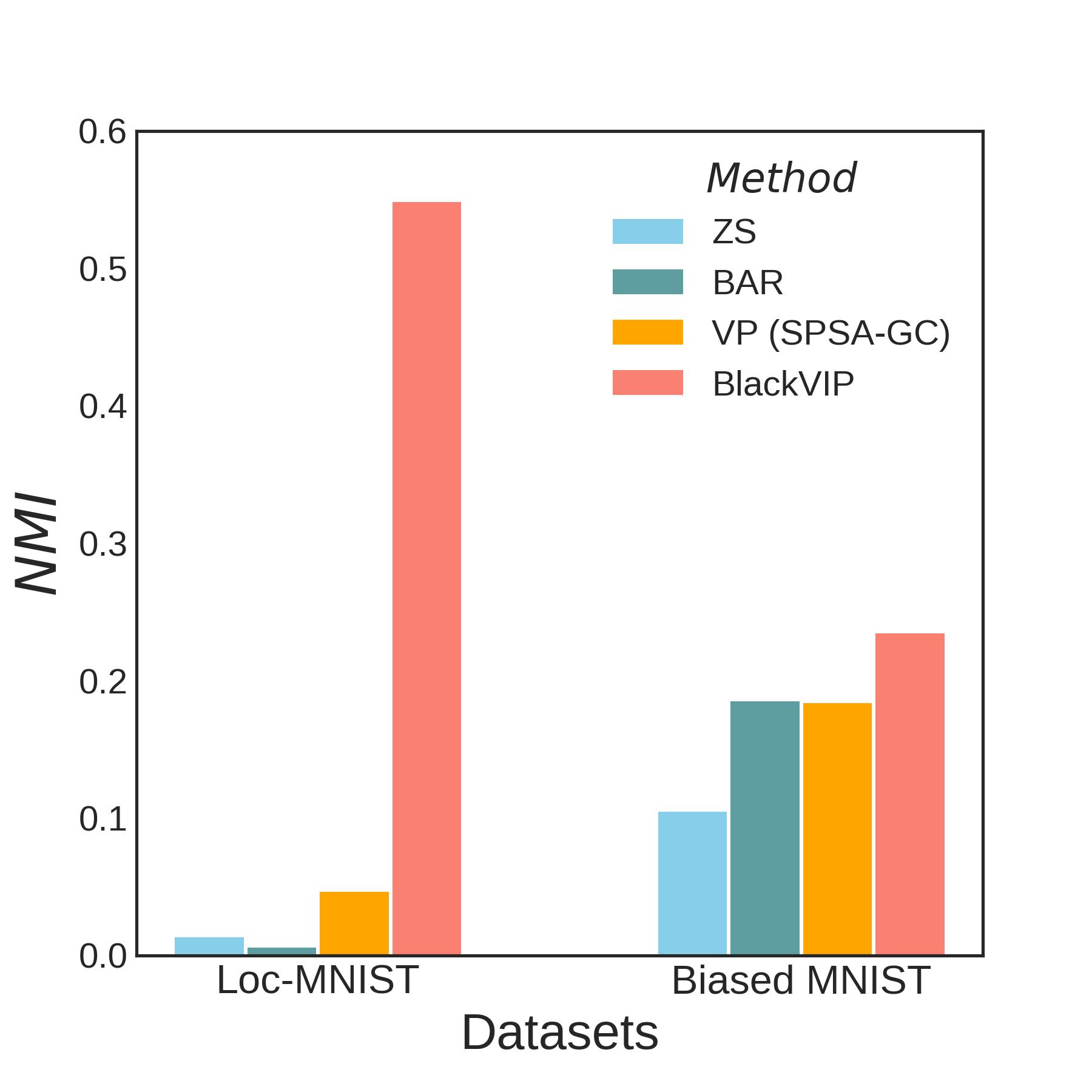}
     \end{subfigure}
    \caption{(Left) t-SNE \cite{van2008visualizing} of prompted images' embedding on Biased MNIST and Loc-MNIST. (right) Normalized Mutual Information (NMI) \cite{loedeman2022prompt} score of learned embedding.}
    \label{fig:synthetic_quantative}
\vspace{-1em}
\end{figure}

\begin{table*}[!t]
\centering
\small
\caption{Results on synthetic datasets. BlackVIP shows robust performance under distribution/object-location shift.}
\begin{tabular}[!t]{@{}l|cccc|cccc@{}}
\toprule
& \multicolumn{4}{c}{Biased MNIST} & \multicolumn{4}{c}{Loc-MNIST} \\ \cmidrule(r){2-9} 
\multirow{2}{*}{Method} & \multicolumn{2}{c}{16-Shot} & \multicolumn{2}{c}{32-Shot} & \multicolumn{2}{c}{16-Shot} & \multicolumn{2}{c}{32-Shot} \\ 
 & $\rho=0.8$ & $\rho=0.9$ & $\rho=0.8$ & $\rho=0.9$ & 1:1 & 1:4 & 1:1 & 1:4 \\ \cmidrule(r){1-9} 
VP (white-box) & {\color[HTML]{D1D1D1}57.92} & {\color[HTML]{D1D1D1}43.55} & {\color[HTML]{D1D1D1}69.65} & {\color[HTML]{D1D1D1}42.91} & {\color[HTML]{D1D1D1}86.79} & {\color[HTML]{D1D1D1}86.54} & {\color[HTML]{D1D1D1}90
.18} & {\color[HTML]{D1D1D1}92.09} \\ \cmidrule(r){1-9}
ZS & 37.56 & 37.25 & 37.56 & 37.25 & 29.70 & 22.70 & 29.70 & 22.70\\
BAR & 53.25 & 53.07 & 53.93 & 53.30 & 33.98 & 26.05 & 34.73 & 27.72 \\
VP w/ SPSA-GC & 60.34 & 53.86 & 59.58 & 51.88 & 16.21 & 25.68 & 18.43 & 30.13 \\
BlackVIP & \textbf{66.21} & \textbf{62.47} & \textbf{65.19} & \textbf{64.47} & \textbf{69.08} & \textbf{60.86} & \textbf{76.97} & \textbf{67.97}\\
\bottomrule
\end{tabular}
\label{tab:syn_results} 
\end{table*}

\paragraph{Robustness on Distribution Shift}
Next, we evaluate our method on Biased MNIST \cite{bahng2020learning} to investigate the robustness of BlackVIP's input-dependent automatic prompt design under distribution shift. Biased MNIST is a modified version of MNIST \cite{lecun1998gradient}, constructed to validate a model's generalization ability under color bias shift. At train-time, each digit has a unique preassigned background color that strongly correlates with the label. The degree of correlation is determined by the value $\rho \in [0, 1]$, and the correlation ratio is reversed as 1-$\rho$ at test-time. Results are summarized in Tab. \ref{tab:syn_results} (left) and Fig. \ref{fig:synthetic_quantative}, respectively. In this setup, BlackVIP remarkably outperforms others (even white-box VP), and the performance gap goes larger under the stronger correlation. This means our input-dependent image-shaped prompts can be beneficial in \textit{domain generalization} settings.
\vspace{-2em}
\paragraph{Robustness on Object Location Shift}
We expect that BlackVIP adopts input-dependent image-shaped prompts, so does be still robust even if the object is not always located in the center of the image. To validate this, we create a variant of the MNIST, Loc-MNIST, by putting a real target digit on the four edges and an arbitrary fake digit in the center of the black blank image. The location of the target digit and the class of the fake digit are chosen randomly. We further consider a more challenging setup in that the fake digit is four times larger (1:4) than the real one. We summarize the results in Tab. \ref{tab:syn_results} (right) and Fig. \ref{fig:synthetic_quantative}, respectively. Compared to input-independent frame-shaped prompting (BAR and VP), BlackVIP achieves significantly better performance which proves the superiority of the Coordinator's prompt design.

\subsection{Few-shot Transfer Learning on Benchmarks}\label{sec:few}
\begin{figure}[htbp]
    \footnotesize
     \centering
     \includegraphics[width=0.9\linewidth]{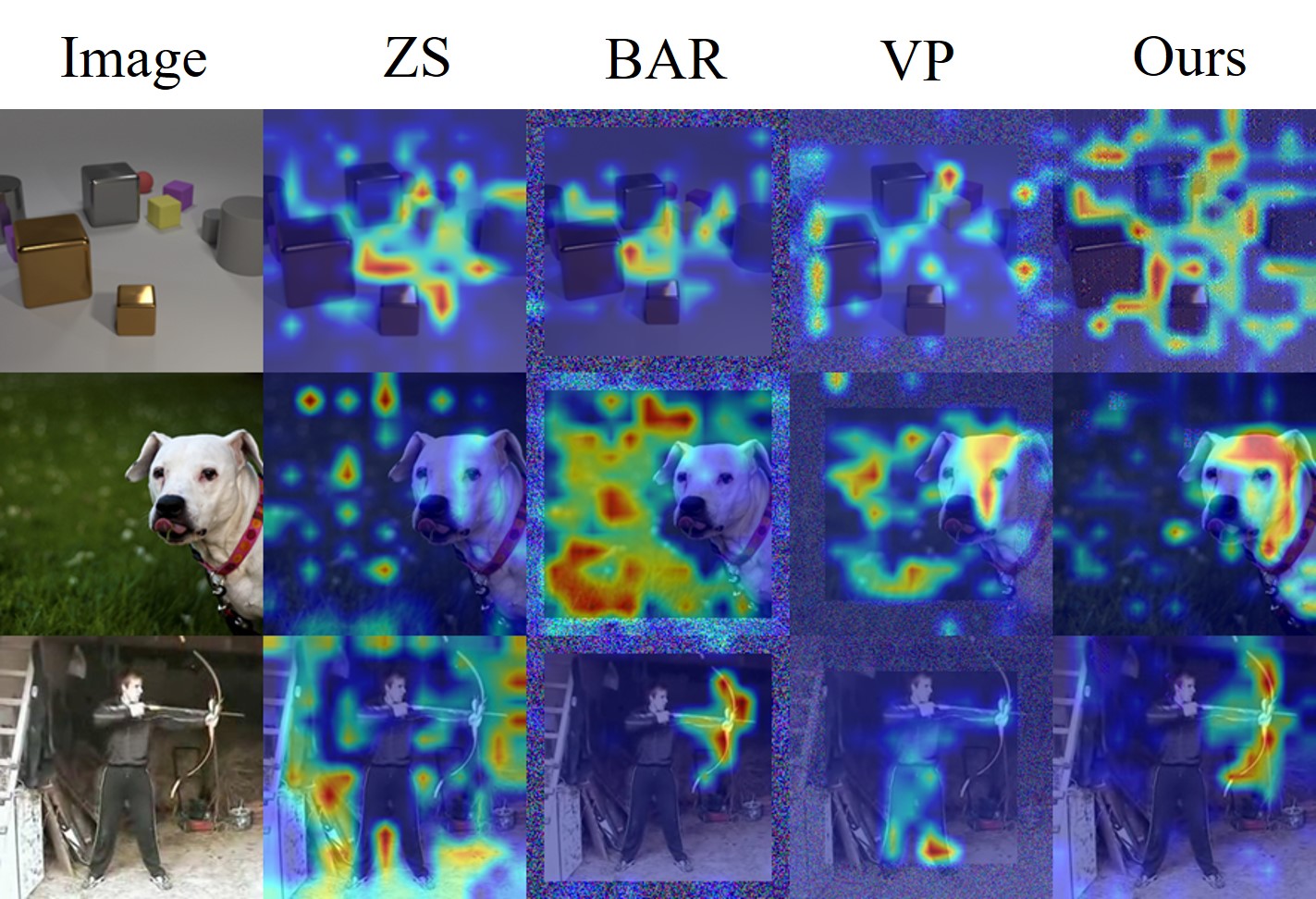} %
    \caption{Grad-CAM analysis on CLVER, Pets, and UCF101.}
    \label{fig:gcam_main}
\end{figure}

We consider the 14 few-shot benchmark datasets following \cite{zhou2022learning, zhou2022conditional, bahng2022visual}. As shown in Tab. \ref{tab:fs14}, while BAR and VP undergo large performance variations across 14 datasets, BlackVIP boasts consistently high performance (i.e., improves the zero-shot performance on 13 over 14 datasets). Specifically, BAR shows promising results on the tasks that require understanding coarse semantics (DTD \cite{6909856}, EuroSAT \cite{helber2019eurosat}, and RESISC \cite{7891544}), but fails to show competitiveness on CLEVR \cite{johnson2017clevr} that requires visual reasoning (counting objects) by capturing the overall image semantics. Meanwhile, BlackVIP performs well across various tasks by extending or limiting attention of frozen PTM (Fig. \ref{fig:gcam_main}), which denotes BlackVIP is a high-capability prompt learner that robustly adapts the PTM to diverse downstream tasks.  

\begin{figure}[htbp]
\centering
\includegraphics[width=\linewidth]{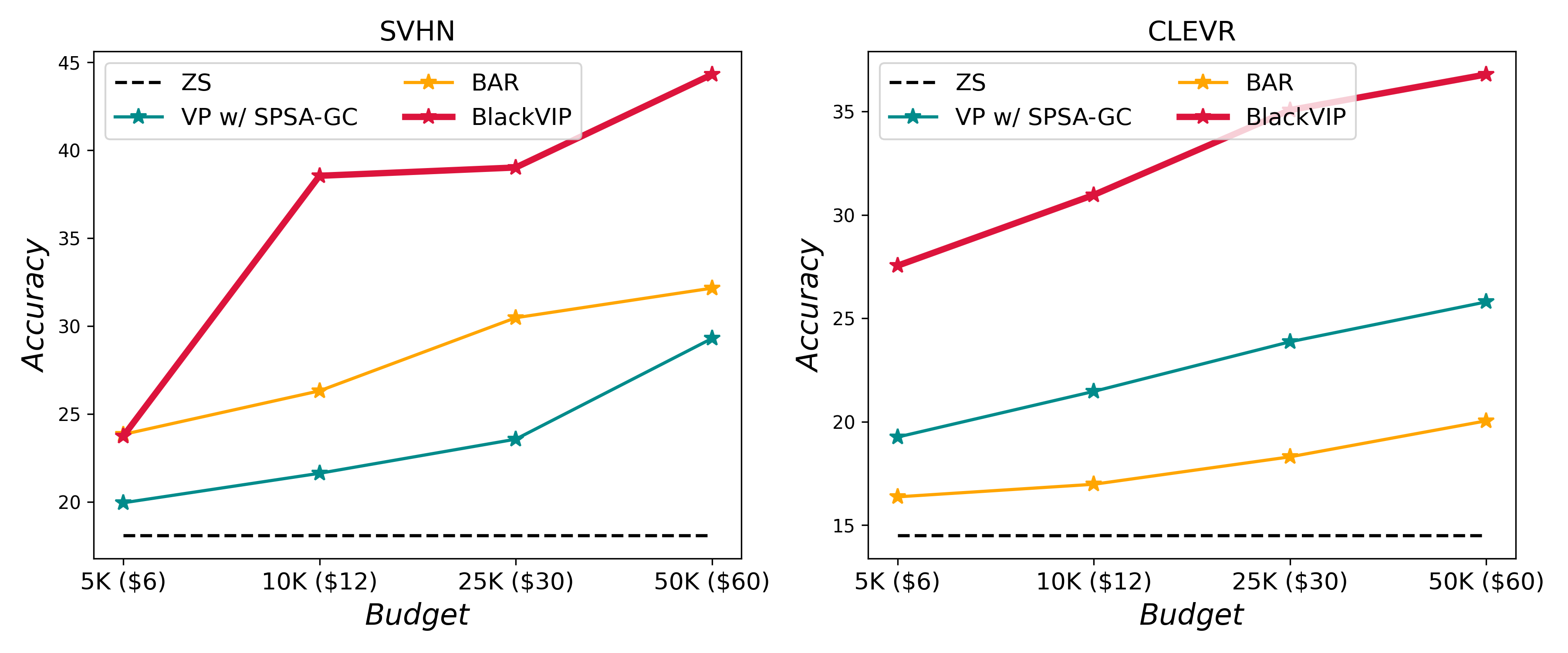}
\captionof{figure}{Query efficiency. (x-axis) A number of queries and cost for achieving (y-axis) corresponding performance.}
\label{fig:query_efficiency}
\end{figure}

Practically, BlackVIP has three major advantages: 1) it only requires the 9K learnable parameters (see Tab. \ref{tab:memory}), while BAR and VP require 37K and 69K parameters. 2) It greatly reduces the peak memory allocation compared to white-box transfer learning methods. 3) BlackVIP shows outstanding query efficiency among the prompting methods (see Fig. \ref{fig:query_efficiency}). For instance, by sending just 10K queries with 12 USD (based on Clarifai Vision API), we can improve the performance of a zero-shot model about twice.

\begin{table*}
\caption{Classification accuracy across 14 benchmarks that require natural, specialized, structured, and fine-grained visual recognition. BlackVIP shows outstanding results among input-space prompting methods. \textit{Win} means the number of datasets that each method beats the zero-shot performance. Grays are the results of white-box learning. All experiments are done in 16 shots with three repeated runs.}
\vspace{-1.5em}
\label{tab:fs14}
\begin{center}
\begin{small}
\resizebox{\textwidth}{!}{\begin{tabular}{@{}l|cccccccccccccc|c|c@{}}
\toprule
Method & Caltech & Pets  & Cars & Flowers & Food & Aircraft & SUN & DTD & SVHN & EuroSAT & RESISC & CLEVR & UCF & IN & \textit{Avg.} & \textit{Win} \\ \midrule
VP (white-box)  & {\color[HTML]{D1D1D1} 94.2} & {\color[HTML]{D1D1D1} 90.2} & {\color[HTML]{D1D1D1} 66.9} & {\color[HTML]{D1D1D1} 86.9} & {\color[HTML]{D1D1D1}81.8} & {\color[HTML]{D1D1D1} 31.8} & {\color[HTML]{D1D1D1} 67.1} & {\color[HTML]{D1D1D1} 61.9} & {\color[HTML]{D1D1D1}60.4} & {\color[HTML]{D1D1D1} 90.8} & {\color[HTML]{D1D1D1}81.4} & {\color[HTML]{D1D1D1}40.8} & {\color[HTML]{D1D1D1} 74.2} & {\color[HTML]{D1D1D1} 67.4 } & {\color[HTML]{D1D1D1} 71.1} & 13\\ \midrule
ZS & 92.9 & 89.1 & 65.2 & \textbf{71.3} & 86.1 & 24.8 & 62.6 & 44.7 & 18.1 & 47.9 & 57.8 & 14.5 & 66.8 & 66.7 & 57.6 & - \\
BAR & \textbf{93.8} & 88.6 & 63.0 & 71.2 & 84.5 & 24.5 & 62.4 & \textbf{47.0} & 34.9 & \textbf{77.2} & \textbf{65.3} & 18.7 & 64.2 & 64.6 & 61.4 & 6 \\
VP w/ SPSA-GC & 89.4 & 87.1 & 56.6 & 67.0 & 80.4 & 23.8 & 61.2 & 44.5 & 29.3 & 70.9 & 61.3 & 25.8 & 64.6 & 62.3 & 58.8 & 4 \\
BlackVIP  & 93.7 & \textbf{89.7} & \textbf{65.6} & 70.6 & \textbf{86.6} & \textbf{25.0} & \textbf{64.7} & 45.2 &  \textbf{44.3} & 73.1 & 64.5 & \textbf{36.8} & \textbf{69.1} & \textbf{67.1} & \textbf{64.0} & \textbf{13} \\  \bottomrule
\end{tabular}}
\end{small}
\end{center}
\end{table*}

\begin{table}
\centering
\footnotesize
\captionof{table}{Train-time peak memory allocation (Peak Memory) and the number of learnable parameters (Params) on ImageNet.}
\label{tab:memory}
\begin{tabular}{@{}lcccc@{}}
\toprule
\multirow{2}{*}{Method} & \multicolumn{2}{c}{Peak Memory (MB)} & \multicolumn{2}{c}{Params} \\ \cmidrule(l){2-5}
 & \multicolumn{1}{c}{ViT-B} & \multicolumn{1}{c}{ViT-L} & ViT-B & ViT-L \\ \midrule
FT (white-box) & 21,655 & 76,635 & 86M  & 304M  \\ 
LP (white-box) & \textbf{1,587} & 3,294 & 513K & 769K  \\
VP (white-box) & 11,937 & 44,560 & 69K & 69K \\
\midrule
BAR & 1,649 & 3,352 & 37K  & 37K \\
VP w/ SPSA-GC & 1,665 & 3,369 & 69K & 69K  \\
BlackVIP & 2,428 & \textbf{3,260} & \textbf{9K}  & \textbf{9K}  \\ \bottomrule
\end{tabular}
\end{table}

\subsection{Ablation Study} \label{sec:ablation}
In this section, we provide two additional results for our BlackVIP: 1) we validate whether BlackVIP can achieve superior performance across four target backbones and two encoders of Coordinator. 2) we present the ablation study about pre-trained weights and optimization algorithms.

\paragraph{Architectures}
To study the versatility of our method, we vary the backbone architecture of the pre-trained target model and the encoder of Coordinator in Tab. \ref{tab:ablation_backbone}. While BAR and the naive application of SPSA-GC on VP fail to improve the zero-shot performance of CNN-based target backbones that lack the global attention of Transformers \cite{vaswani2017attention}, our BlackVIP consistently brings huge performance gains across all the architectures. It implies that BlackVIP is an \textit{architecture-agnostic} approach, which pursues the general adaptation method for high-performing PTMs.

\begin{table}[t!]
\centering
\footnotesize
\caption{Ablation study for backbone architecture. Classification accuracy on EuroSAT across pre-trained target backbone architectures and BlackVIP's Coordinators (SSL encoder backbone).}
\begin{tabular}{@{}lcccc|c@{}}
\toprule
\textbf{Method} & \multicolumn{5}{c}{\textbf{Target Backbone}} \\ 
\multicolumn{1}{l}{} & RN50 & RN101 & ViT-B/32 & ViT-B/16 & \textit{Avg.} \\ \midrule
{ZS} & 37.5 & 32.6 & 45.2 & 40.8 & 48.4 \\
{BAR} &  {\color[HTML]{dc143c}26.9} & {\color[HTML]{008000}33.5} & {\color[HTML]{008000}{70.3}} & {\color[HTML]{008000}\textbf{77.2}} & 52.0 \\
{VP w\ SPSA-GC} & {\color[HTML]{dc143c}34.7}& {\color[HTML]{dc143c}31.2}& {\color[HTML]{008000}\textbf{71.1}}& {\color[HTML]{008000}70.9} & 52.0 \\
 Ours (RN50) & {\color[HTML]{008000}\textbf{51.3}} & {\color[HTML]{008000}{50.8}} & {\color[HTML]{008000}62.9} & {\color[HTML]{008000}68.5} & {58.4} \\
Ours (VIT-B/16) & {\color[HTML]{008000}{48.4}} & {\color[HTML]{008000}\textbf{51.3}}& {\color[HTML]{008000}67.9} & {\color[HTML]{008000}{73.1}} & \textbf{60.2} \\ \bottomrule
\end{tabular} \label{tab:ablation_backbone}
\end{table}

\begin{table}[htbp]
\small
\centering
\caption{Different Coordinator weights with SPSA variants. Mean classification accuracy of three repeated runs on EuroSAT}
\begin{tabular}{@{}c|cc@{}}
\toprule
Encoder Type & Optim. & Acc. \\ \midrule
\multicolumn{2}{c}{Zero-Shot} & 47.9 \\ \midrule
\textit{scratch} & SPSA & 49.6 \\
\textit{scratch} & SPSA-GC & 49.5 \\
\textit{Sup. pre-trained} & SPSA & 59.4 \\
\textit{Sup. pre-trained} & SPSA-GC & 65.2 \\
\textit{SSL pre-tained} & SPSA & 69.4 \\ \midrule
\multicolumn{2}{c}{\textbf{BlackVIP} (\textit{SSL pre-trained} with SPSA-GC)} & \textbf{73.1} \\ \bottomrule
\end{tabular} \label{tab:ablation}
\end{table}

\paragraph{Coordinator weights and ZOO algorithms} BlackVIP adopts the encoder-decoder structure to efficiently generate the input-dependent image-shaped prompts. We exploit an SSL pre-trained encoder while we plug the randomly initialized extremely lightweight decoder. From the design philosophy of BlackVIP, we expect that a pre-trained encoder extracts the rich semantic features of the given image, including the spatial features, and the decoder utilizes the features to produce a spatially and semantically structured prompt tailored to the input. We conjecture that an SSL pre-trained encoder is desirable to capture the demanding diverse semantics instead of a supervised one learned from pre-defined labels. Therefore, for Coordinator, BlackVIP adopts an SSL encoder (i.e., Masked Auto-Encoder \cite{he2022masked}). Tab. \ref{tab:ablation} confirms that the SSL encoder outperforms the supervised pre-trained or randomly initialized encoder (scratch). Besides, SPSA-GC improves the 3.7\% accuracy than SPSA, from 69.4 to 73.1. It denotes that approximated gradients by our SPSA-GC are more accurate than the original SPSA.

\section{Conclusion}
We pioneer \textit{black-box visual prompting} for the realistic and robust adaptation of pre-trained models. We propose BlackVIP, which reparameterizes the input-space prompt as a conditional generative network Coordinator and equips our new ZOO algorithm, SPSA-GC, rather than backpropagation. BlackVIP does not require any accessibility on model architecture or parameters and efficiently adapts the pre-trained model to targeted downstream tasks. Extensive empirical results show that BlackVIP consistently improves the performance over baseline methods on few-shot adaptation, distribution shift, and object-location shift with minimal parameters, memory capacity, API queries, and cost.

\section*{Acknowledgement}
This work was supported by a grant from National Research Foundation of Korea(2022R1A4A3033874, 2021R1F1A1060117, 2020R1C1C1008726, 2017R1D1A1 B05028565), and supported by Institute of Information \& Communications Technology Planning \& Evaluation grant funded by the Korea government (No.2021-0-02067)

{\small
\bibliographystyle{ieee_fullname}
\bibliography{main}
}

\newpage

\section*{A. Experimental Setting}
\label{app:b_exp_setup}
\subsection*{A.1. Datasets}
\label{app:b_dataset}
\begin{figure}[!htp]
     \centering
     \begin{subfigure}[b]{0.5\textwidth}
         \centering
        \includegraphics[width=0.95\textwidth]{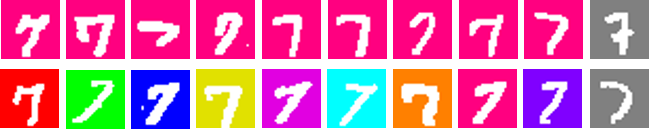}
         \caption{Examples of $y=7$ subset in Biased-MNIST \cite{bahng2020learning} with $\rho=0.9$. (Top) the train set is constructed with the spurious correlation between the background color and digit class (e.g., $y=7$ occurs $90\%$ with pink background and $10\%$ with other random colors in this case). (Bottom) the test set is constructed with a reversed correlation to that of the train set (e.g., $y=7$ occurs $10\%$ with pink background and $90\%$ with other random colors in this case).}
         \label{fig:a_bmnist}
     \end{subfigure}
     \begin{subfigure}[b]{0.5\textwidth}
         \centering
        \includegraphics[width=0.95\textwidth]{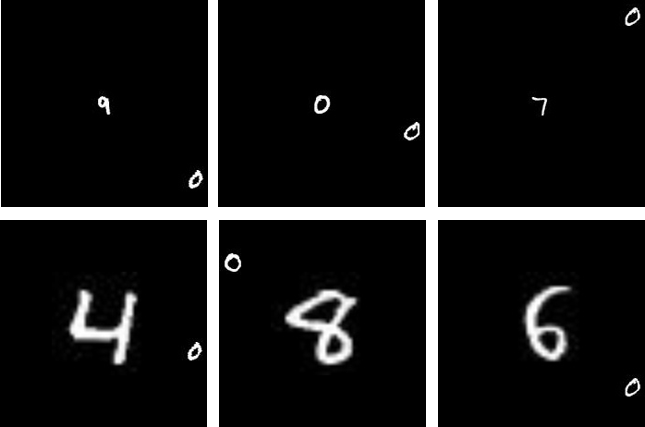}
         \caption{Examples of Loc-MNIST dataset. The real digit from MNIST is located in the outer area, while the fake digit from another random MNIST image is placed in the center of the image. (Top) the case where the size ratio of the real digit to the fake digit is 1:1, and (Bottom) 1:4.}
         \label{fig:a_lmnist}
     \end{subfigure}
     \caption{Examples of two synthetic datasets. (a) Biased MNIST and (b) Loc-MNIST.}
     \label{fig:a_synthetic_examples}
\end{figure}

\paragraph{Synthetic Datasets}
Our BlackVIP generates the input-dependent image-size visual prompt which covers the whole image region, so we expect that this flexible prompt design can improve some kind of robustness as well as general recognition capability: (1) To evaluate the robustness on distribution shift (i.e., domain generalization), we consider Biased-MNIST \cite{bahng2020learning} dataset. (2) To evaluate the robustness on adversarial noise and location-agnostic recognition capacity, we create a variant of the MNIST dataset called Loc-MNIST. Examples of these two datasets are provided in Figure \ref{fig:a_synthetic_examples}.

\textbf{Biased MNIST} is a modified version of MNIST \cite{lecun1998gradient} where the biases reside in the background colors of the images of each digit. At train time, each digit has a unique preassigned background color that strongly correlates with the label. The degree of correlation is determined by the value $\rho \in [0, 1]$, such that $(100 \times \rho) \%$ of the images that belong to the same digit have the preassigned color of that digit as their background color, and the rest are uniformly assigned to have any of the other colors as their background color. At test time, we reverse the ratio so that  $(100 \times (1 - \rho)) \%$ of the images now have the preassigned color as their background color and vice versa to evaluate the model’s dependency on superficial features such as the color of the background that a digit is located on. We prepare the following two environments 1) easy: $\rho=0.8$ and 2) hard: $\rho=0.9$.

On the given black blank image with $224\times224$ resolution, i.e., zero's array, \textbf{Loc-MNIST} puts an original target digit image from MNIST that has $28\times28$ resolution on the edge-side (e.g., 0$\sim$27 or 196$\sim$223 for one of vertical or horizontal side and 0$\sim$223 for another side) and puts a random fake digit (also from the MNIST dataset) on the center. The location of the target digit in the edge and the class of fake digit are chosen randomly with uniform probability. A synthetic image is created one by one for each original MNIST image. We prepare the following two environments 1) easy: the scale of the target and the fake digit is the same, i.e., 1:1, and 2) hard: the fake digit is four times larger than the original digit, i.e., 1:4.

For consistency, we perform the experiments on these two datasets with a few-shot evaluation protocol. To construct a train set, we randomly sample a subset (K-shot) of the created images for each class and use the whole test set.

\paragraph{Datasets}
To extensively evaluate the effectiveness of our proposed method and baseline approaches, we measure performance across the following 14 datasets that are widely used for transfer learning benchmark: Caltech101 \cite{1384978}, OxfordPets \cite{6248092}, StanfordCars \cite{6755945}, Flowers102 \cite{4756141}, Food101 \cite{10.1007/978-3-319-10599-4_29}, FGVCAircraft \cite{maji2013fine}, SUN397 \cite{5539970}, DTD \cite{6909856}, SVHN \cite{37648}, EuroSAT \cite{helber2019eurosat}, Resisc45 \cite{7891544}, CLEVR \cite{johnson2017clevr}, UCF101 \cite{soomro2012ucf101}, and ImageNet (IN) \cite{5206848}. Note that these 14 datasets cover diverse visual domains, and they require understanding various visual semantics like scenes, actions, fine-grained categories, textures, satellite imagery, digits, the number of objects, and the recognition of generic objects.

Following the protocol in \cite{zhou2022learning, zhou2022conditional}, we conduct a few-shot evaluation for all datasets: 16-shot for the train set, 4-shot for the validation set, and the whole test set. We use the few-shot split by \cite{zhou2022learning} for each dataset those are also used in \cite{zhou2022learning}, while for Resisc45 and CLEVR, we randomly select the 16-shot and 4-shot samples for training and validation dataset, respectively.

\subsection*{A.2. Backbone Model}
\label{app:b_backbone}
In this work, we aim at the robust adaptation of pre-trained models on diverse downstream tasks. For these pre-trained models, all experiments in this paper are done with the off-the-shelf vision-language model CLIP \cite{radford2021learning}, and we adopt the ViT-B/16 for image encoder backbone architecture by default. During the adaptation (training) phase, the entire components of the pre-trained model are frozen without any architectural modification, and we only manage and optimize the learnable module Coordinator from the outside of the pre-trained model.

While input space visual prompting allows it to be applied to not only VLM, but also any other vision models like CNNs and ViTs, it requires the user to define the output space mapping, which maps the output prediction category set of a pre-trained task to a new downstream category set \cite{elsayed2018adversarial, tsai2020transfer, bahng2022visual}. This is another non-trivial problem. Therefore, we limit our focus to only the VLM that can dynamically build the task-specific head from manual text template \cite{radford2021learning, jia2021scaling} so that free from defining output space mapping.

\subsection*{A.3. Baseline Methods}
\label{app:b_baseline}
\paragraph{CLIP Zero-Shot (ZS)} 
CLIP \cite{radford2021learning} is one of the most popular vision-language zero-shot models that is widely exploited for classification, detection, segmentation, and other vision or vision-language tasks. Based on its well-aligned vision-language joint embedding space, the zero-shot classification can be performed with a manual text prompt (also called template) of each pre-defined class category. In this paper, we are mainly aiming to improve the CLIP's strong zero-shot performance in the few-shot adaptation setting. 

\paragraph{BAR} Black-Box Adversarial Reprogramming (BAR) \cite{tsai2020transfer} was proposed for efficient transfer learning of pre-trained model to the medical image domain. Different from the previous works on Adversarial Reprogramming (AR), BAR exploits the perturbation-vulnerability of neural networks for \textit{adaptation} purpose rather than attack. By optimizing the frame-shaped learnable program, which embeds a downstream target image inside of that, BAR steers the ImageNet pre-trained model to classify the specialized medical images. Moreover, BAR adopts the zeroth-order optimizer (ZOO), Randomized Gradient-Free (RGF) \cite{10.1007/s10208-015-9296-2} minimization algorithm for black-box transfer learning to broaden its applications.

When the resolution of the downstream input image is over that of the pre-training phase, Tsai et al. \cite{tsai2020transfer} set the embedded target image size for $64\times64$ resolution in the $299\times299$-size learnable program by default. However, we observe that such a heavy-pad thin-image design of prompt degrade the performance significantly, so we tune the resolution of the embedded image and set $194\times194$.

\paragraph{VP}
Similarly, Visual Prompting (VP) aims at adapting a pre-trained model to downstream tasks via learning input space visual prompts. Among some candidates for prompt designs, Bahng et al. \cite{bahng2022visual} adopt the padding-style prompt so that realized prompts look like the frame-shape program of ARs. VP learns a universal visual prompt per each downstream task, and it just adds to all of the images in a task. Unlike the AR methods or our BlackVIP, the range of prompted images is unbounded. Following \cite{bahng2022visual}, we use the padding-style prompt, which is 30-pixel sized for each side by default.

While VP optimizes the parameters in the input space, it relies on a first-order optimization algorithm that uses the true gradient of entire model parameters, and we establish the performance of VP as an upper bound for other input space black-box optimization approaches, including BlackVIP. Additionally, by replacing the first-order algorithm with zeroth-order counterparts, we build two new baselines \textbf{VP w/ SPSA} and \textbf{VP w/ SPSA-GC} on our extensive experiments. These two methods confirm the effectiveness of our new components \textit{Coordinator} and SPSA-GC.

\paragraph{Discussion}
Although BAR, VP, and BlackVIP share the generic goal: efficient transfer learning of pre-trained models via input-space optimization, there are several significant differences. (1) We propose a novel prompt design that is automatically formed in an input-dependent manner rather than the frame-shaped manual design of the input-independent prompt (or program) of VP (or BAR). (2) While VP relies on first-order algorithms and BAR adopts the RGF, we utilize the new variants of SPSA \cite{119632}, SPSA-GC, which is enhanced with a proper modification in the parameter update rule. (3) Contrary to the medical imaging-only validation in BAR, based on the above two technical difference, BlackVIP successfully adapt the pre-trained model to diverse data domains (described in Section B.1.).

\subsection*{A.4. Implementation Details}
\label{app:b_implementation}
\paragraph{Architecture}
For the fixed text prompt design of each dataset those are shared across all baseline methods and BlackVIP, we use the same templates provided by  \cite{bahng2022visual} for SVHN, CLEVR, and Resisc45, and \cite{zhou2022learning} for remaining 11 datasets. For the frozen feature extractor (encoder) part of our \textit{Coodinator}, we use the ImageNet pre-trained \texttt{vit-mae-base} checkpoint\footnote{\url{https://huggingface.co/docs/transformers/model_doc/vit_mae}} from the HuggingFace. The output shape of the encoder is $N\times768$, where $N$ is the number of instances in the batch. We design the decoder based on \textit{depth-wise separable convolution} (DSC) layer \cite{chollet2017xception} for parameter efficiency. Specifically, we build a block of [\texttt{NORM-ACT-CONV}] and stack it five times. The \texttt{NORM} and \texttt{ACT} denote Batch Normalization and Gaussian Error Linear Unit, respectively. The \texttt{CONV} operation of the first four blocks is DSC, and the last one is a standard convolutional layer. Our implementation code is enclosed in \texttt{.zip} file.

To satisfy a fully convolutional design without loss of expressiveness, tensors that are fed into the decoder must be shaped in a 3D feature map. For this, we additionally govern a task-specific single continuous vector $\phi_{t}$ (called \textit{prompt trigger vector}), which is concatenated with the output feature vector of encoder leading the appropriate size of 1d vector for reshaping to 3d tensor. In this work, we set the dimension of the prompt trigger vector to 800, resulting in 1568 dimensions of concatenated vector that can be reshaped to $32\times7\times7$ shaped 3D tensor. The prompt trigger is shared across all instances for a given task.

\paragraph{Optimization and other configurations}
For a stable approximation of gradient in practice, ZOO algorithms repeat the gradient estimation step for several times and use the mean of those estimates as a final approximation of the gradient. Usually, the approximation quality is proportional to the number of these repeats. We set this repeat as five times for all baselines that use ZOO. 

Besides the learning rate and learning rate schedule parameters, ZOO algorithms have some additional algorithm-specific hyperparameters needed to be tuned. For RGF, these are the standard deviation of a random gaussian vector and a smoothing parameter, and for SPSA, these are the perturbation magnitude and its decaying factor. We provide the search range of each hyperparameter in Table \ref{tab:a_hyparam}. The search range for algorithm-specific parameters is based on the proposal of authors of SPSA \cite{10.5555/773301} and BAR \cite{tsai2020transfer}. Moreover, among the valid perturbation distributions of SPSA, we adopt the Segmented Uniform $[-1.0, -0.5] \cup [0.5, 1.0]$.

The learning objective is a cross-entropy loss for VP and BlackVIP and focal loss for BAR (following \cite{tsai2020transfer}). For all black-box approaches, the batch size is set to 128 across all datasets. Except for the SUN397 (1,000), StanfordCars (2,500), and ImageNet (500), we optimize all methods during 5,000 epochs for convergence. Note that the input space visual prompting with first-order algorithm already requires sufficiently large iterations, e.g., 1,000 epoch \cite{bahng2022visual} with full dataset, and ZOO demands much more iterations due to the lack of gradient information. 

\subsection*{A.5. Hyperparameter Sweep}
\label{app:b_hyperparameter}
In this section, we provide the hyperparameter search range of each algorithm, summarized in Table \ref{tab:a_hyparam}.

\begin{table}[htbp]
\centering
\caption{Hyperparameter sweep. Large LR (learning rate) of BAR and VP is based on \cite{bahng2022visual} to directly optimize pixel values rather than the neural network's weights. PM denotes perturbation scale, $c_{i}$.}
\label{tab:a_hyparam}
\resizebox{0.5\textwidth}{!}{\begin{tabular}{@{}lcc@{}}
\toprule
Hyperparameter & Algorithm & Search Range \\ \midrule
initial LR & BAR, VP & $\{$40.0, 20.0, 10.0, 5.0, 1.0$\}$ \\
initial LR ($a_{1}$) & BlackVIP & $\{$1.0, 0.1, 0.01, 0.005$\}$ \\
min LR & BAR & $\{$0.1, 0.01, 0.001$\}$ \\
decaying step & BAR & $\{$0.9, 0.5, 0.1$\}$ \\
LR decaying factor & VP, BlackVIP & $\{$0.6, 0.5, 0.4, 0.3$\}$ \\
initial PM ($c_{1}$) & BlackVIP & $\{$0.01, 0.005, 0.001$\}$ \\
PM decaying factor & BlackVIP & $\{$0.2, 0.1$\}$ \\
std. of perturbation & BAR & $\{$1.0, 0.5$\}$ \\
smoothing & BAR & $\{$0.1, 0.01, 0.001$\}$ \\ 
gradient smoothing & VP, BlackVIP & $\{$0.9, 0.7, 0.5, 0.3$\}$ \\ \bottomrule
\end{tabular}}
\end{table}

\section*{B. Detail Description of \textit{Coodinator}}
\label{app:c_model_desc}
On the transfer learning of a pre-trained model which provides no accessibility about any architectural information or actual model parameters, BlackVIP treats this situation with two novel mechanisms: (1) parameter-efficient instance-aware prompt generation network, and (2) stable zeroth-order optimization algorithm that is based on SPSA \cite{119632}. In this section, we provide a detailed description of the first component, Coordinator.

Different from existing works on visual prompting, we reparameterize the input space visual prompt $\phi$ as a neural network, \textit{Coordinator} $h_{\bm\phi}(\cdot)$ that generates an input-dependent visual prompt $h_{\bm\phi}(x)$. Coordinator is composed with encoder $f(\cdot)$, decoder $g_{\phi_{d}}(\cdot)$ and task-specific learnable vector $\phi_{t}$. The encoder is used for extracting instance-specific latent feature vector $z_{x}=f(x)$ contributing to the construction of the optimal input space visual prompt for each instance. Because our goal in this work is the broad utilization of pre-trained models on diverse downstream tasks, we adopt a pre-trained encoder network optimized by a self-supervised learning objective, not by a supervised learning objective or scratch network. Specifically, we use the ViT-B/16 weights from the \textit{Masked AutoEncoding} pre-training \cite{he2022masked}. We present the grounds for using the self-supervised learning encoder in the main paper, refer to Sec. 3. During the training phase, this pre-trained encoder part is frozen (not updated) and just acts as a feature extractor. Then, the instance-specific feature vector from the encoder is conveyed to the decoder for a prompt generation. 

Prompt decoder $g_{\phi_{d}}(\cdot)$ is a lightweight convolutional neural network, which has learnable parameters \textbf{less than 10K} by default. Note that the generated prompt has the same shape as the input image, so our prompt covers the entire region of the image, unlike previous visual prompting and reprogramming works applied to the partial region of the image by human-designed.

In addition to the feature vector from the fixed encoder, the decoder also incorporates an additional input which is shared for all instances across the current dataset. The so-called \textit{prompt trigger vector} $\phi_{t}$ is a continuous vector that also contributes to the design of a visual prompt by collaborating with the instance-specific rich feature vector from the encoder. By introducing this prompt trigger vector, the decoder of the Coordinator can enjoy additional information to generate more proper prompts for a given task. Besides, it helps to build the 3D feature map for the decoder's input, which is necessary for designing a parameter-efficient fully convolutional decoder network.

\section*{C. Grad-CAM Analysis}
\label{app:a_qual_anal}
\begin{figure*}
    \centerline{\includegraphics[width=0.76\textwidth]{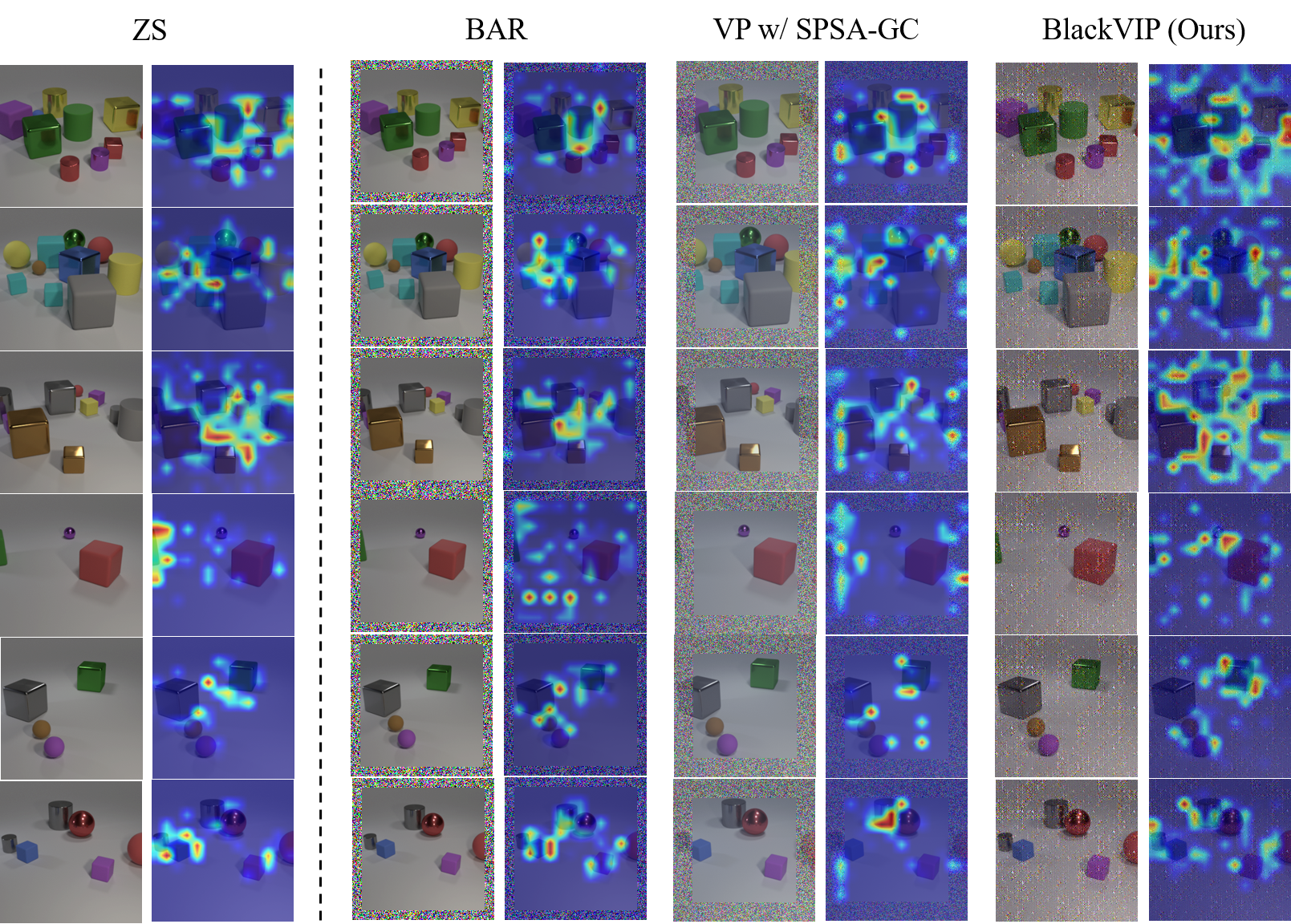}}
    \caption{Grad-CAM on CLEVR. Compared to baseline methods, BlackVIP extends the attention of models to broad areas of the image for effective reasoning on the number of objects.}
	\label{fig:a_gcam_clevr}
\end{figure*}

\begin{figure*} 
    \centerline{\includegraphics[width=0.76\textwidth]{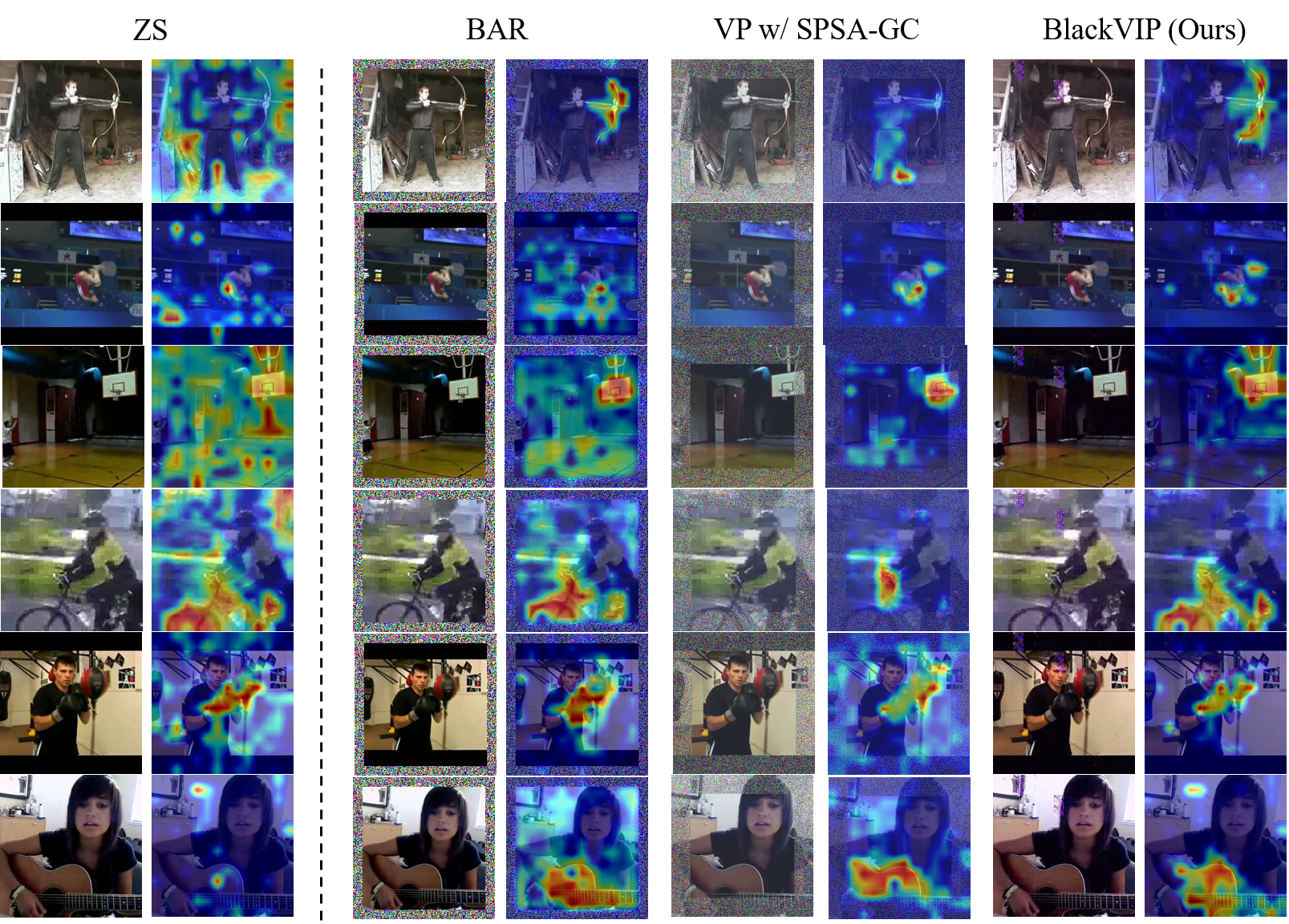}}
    \caption{Grad-CAM on UCF101. Compared to baseline methods, BlackVIP concentrates the attention of models on local areas of the image for effective recognition of the specific actions.}
	\label{fig:a_gcam_ucf}
\end{figure*}

\begin{figure*} 
    \centerline{\includegraphics[width=0.75\textwidth]{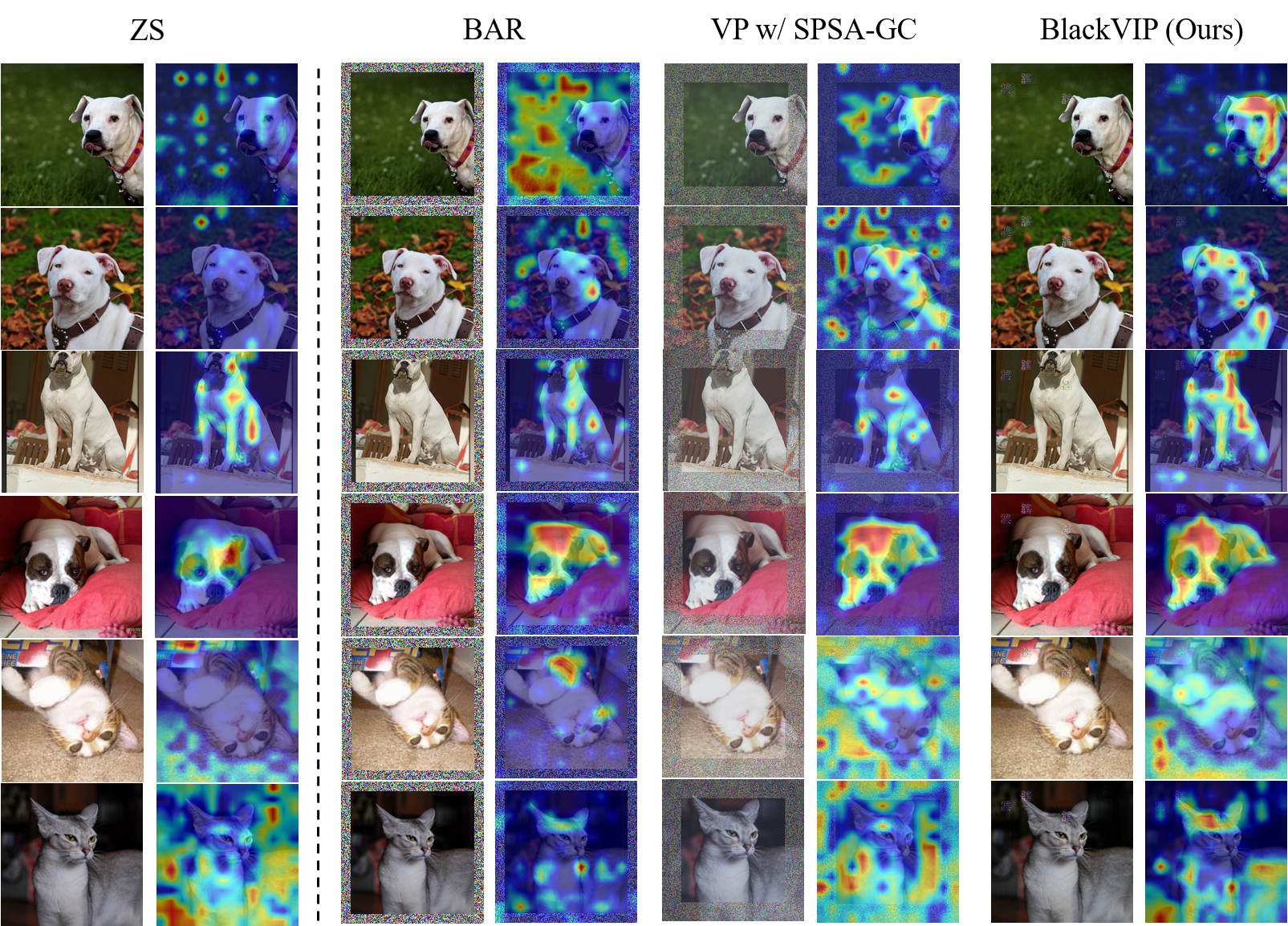}}
    \caption{Grad-CAM on OxfordPets. Compared to baseline methods, BlackVIP effectively adapts the model to focus on the target object rather than spurious features such as the background.}
	\label{fig:a_gcam_pets}
\end{figure*}

\begin{figure*} 
    \centerline{\includegraphics[width=0.76\textwidth]{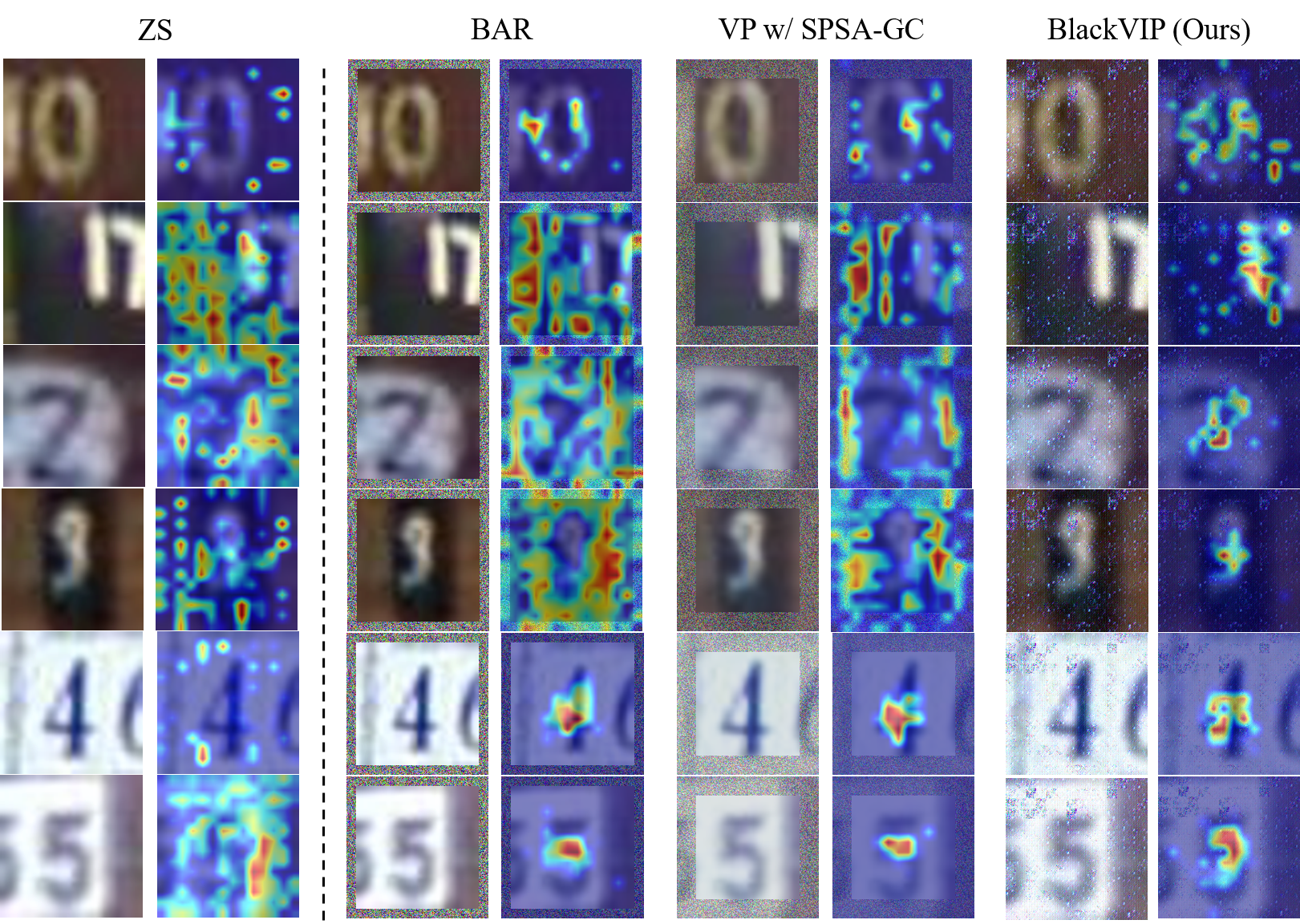}}
    \caption{Grad-CAM on SVHN. Compared to baseline methods, BlackVIP effectively adapts the model to focus on the target digit rather than spurious features such as the background.}
	\label{fig:a_gcam_svhn}
\end{figure*}

\begin{figure*} 
    \centerline{\includegraphics[width=0.76\textwidth]{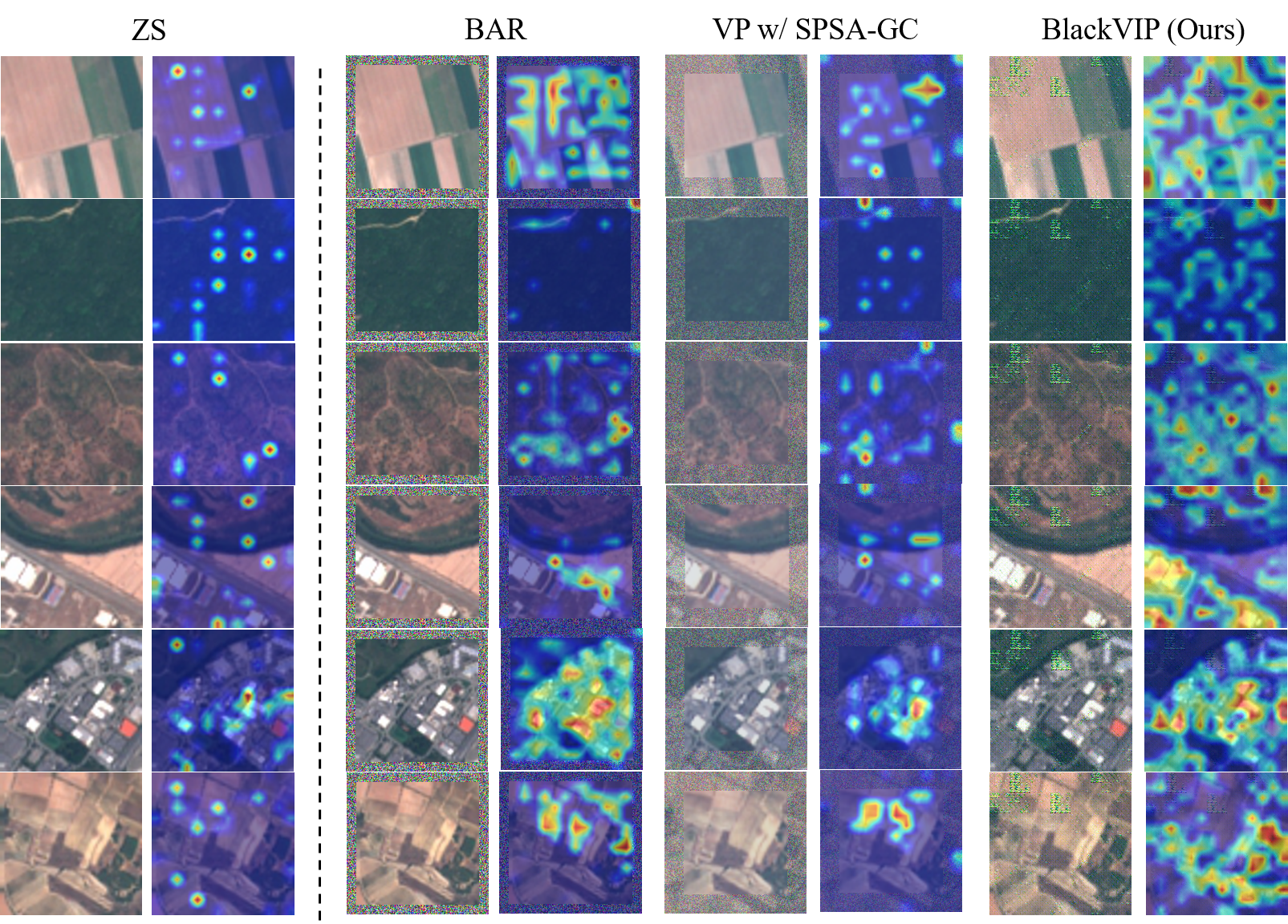}}
    \caption{Grad-CAM on EuroSAT. Compared to baseline methods, BlackVIP extends the attention of models to broad areas of the image for effective classification of satellite imagery.}
	\label{fig:a_gcam_eur}
\end{figure*}

\begin{figure*} 
    \centerline{\includegraphics[width=0.76\textwidth]{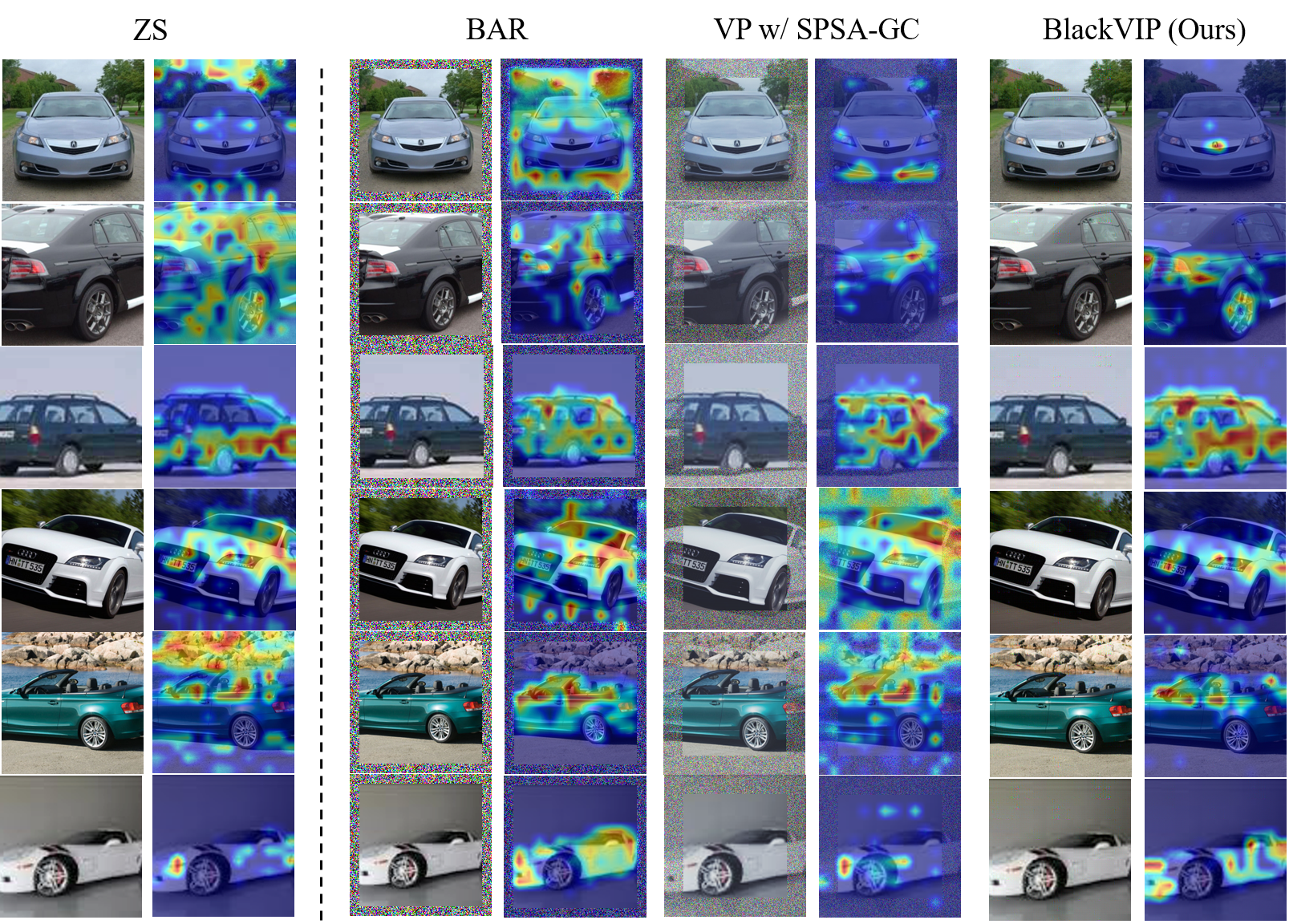}}
    \caption{Grad-CAM on StanfordCars. Compared to baseline methods, BlackVIP concentrates the attention of models on an object or local areas of an image for effective fine-grained classification.}
	\label{fig:a_gcam_cars}
\end{figure*}

\begin{figure*}
    \centerline{\includegraphics[width=0.76\textwidth]{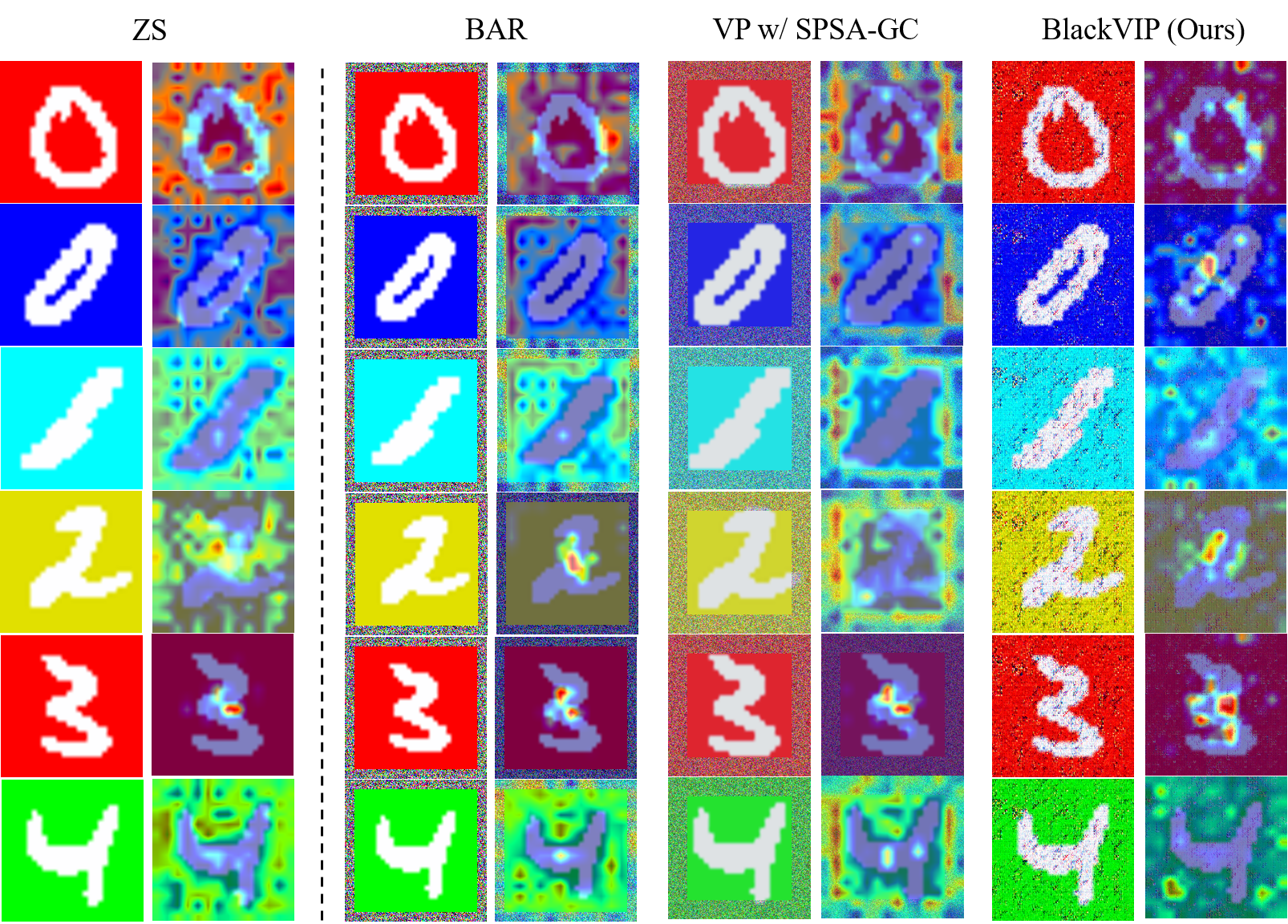}}
    \caption{Grad-CAM on Biased-MNIST. While baseline methods attend to the background rather than digit shape, our BlackVIP can bypass this spurious feature through a widely scattered visual prompt and focus more of the attention on the shape of the digit.}
	\label{fig:a_gcam_bmnist}
\end{figure*}

\begin{figure*} 
    \centerline{\includegraphics[width=0.76\textwidth]{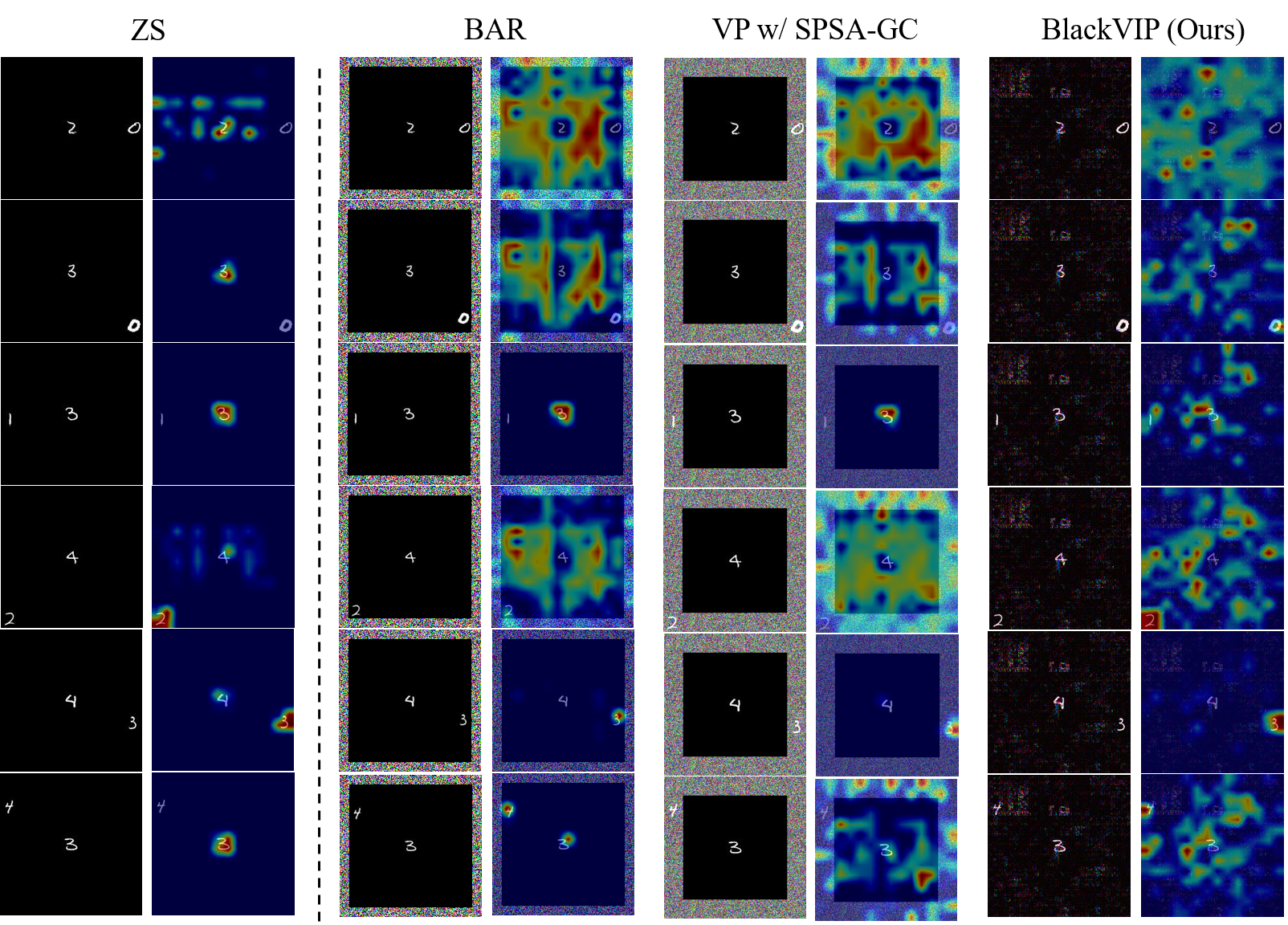}}
    \caption{Grad-CAM on Loc-MNIST. Compared to baseline methods, BlackVIP effectively adapts the model to aim at edge-located true digit corresponding true label rather than the obstructive fake digit in the center of the image.}
	\label{fig:a_gcam_lmnist}
\end{figure*}

To investigate whether visual prompts produced by each method adapt the pre-trained model, we visualize the Grad-CAM \cite{selvaraju2017grad} on the original image and prompted image of the encoder's penultimate layer (Figure \ref{fig:a_gcam_clevr}-\ref{fig:a_gcam_lmnist}). We select eight datasets that represent the diverse image domains and experimental conditions: (\textit{Natural}) OxfordPets and SVHN, (\textit{Specialized}) EuroSAT, (\textit{Structured}) CLEVR-count, (\textit{Action Recognition}) UCF101, (\textit{Fine-Grained}) StanfordCars, (\textit{Synthetic}) Biased-MNIST and Loc-MNIST. Detail descriptions for each dataset are provided in Sec. A.1.

BlackVIP generates diverse prompts to properly adapt the pre-trained model to the targeted data domains. When the task is counting the number of objects (CLEVR) in the entire region of an image, BlackVIP extends the attention of the pre-trained model to the whole view of an image as shown in Figure \ref{fig:a_gcam_clevr}. If the task requires a fine-grained understanding of objects or recognition of actions (UCF101), BlackVIP concentrates the model's attention on class-related regions, as shown in Figure \ref{fig:a_gcam_ucf}.

\newpage
\end{document}